\documentclass{article}

\usepackage{xspace}
\usepackage{graphicx}
\usepackage{subfigure}
\usepackage{booktabs} 
\usepackage[normalem]{ulem}
\usepackage{listings}
\usepackage{xcolor}
\usepackage{amsmath, amssymb, amsthm, mathtools}
\usepackage{tikz}
\usepackage{tikz-3dplot}
\usepackage{algorithm}
\usepackage[noend]{algorithmic}

\usepackage{hyperref}

\usepackage[accepted]{mlsys2021}

\definecolor{myorange}{HTML}{f99353}
\definecolor{myblue}{HTML}{5791e9}

\theoremstyle{definition}

\usetikzlibrary{calc}
\usetikzlibrary{positioning}
\usetikzlibrary{shapes.arrows}
\usetikzlibrary{matrix,arrows.meta,fit}
\newcommand{\figref}[1]{Fig.~\ref{Fi:#1}}
\newcommand{\tabref}[1]{Table~\ref{Ta:#1}}
\newcommand{\secref}[1]{Section~\ref{Se:#1}}

\newcommand{\mypar}[1]{{\bf #1.}}

\newcommand{\mus}{\phantom{ms}\makebox[0pt][r]{\textmu s}}
\newcommand{\dependence}[1]{\mathcal{D}^{#1}}

\newcommand{\deeppoly}{DeepPoly\xspace}
\newcommand{\salsa}{GPUPoly\xspace}
\newcommand{\tool}{\salsa}

\newcommand{\crown}{{CR-IBP}\xspace}
\newcommand{\layerno}{\ell}
\newcommand{\pluseq}{\mathrel{+}\mathrel{\mkern-2mu}=}

\mlsystitlerunning{Scaling Polyhedral Neural Network Verification on GPUs}

\begin{document}

\twocolumn[
\mlsystitle{Scaling Polyhedral Neural Network Verification on GPUs}



\mlsyssetsymbol{equal}{*}

\begin{mlsysauthorlist}
\mlsysauthor{Christoph Müller}{equal,to}
\mlsysauthor{François Serre}{equal,to}
\mlsysauthor{Gagandeep Singh}{goo}
\mlsysauthor{Markus Püschel}{to}
\mlsysauthor{Martin Vechev}{to}
\end{mlsysauthorlist}

\mlsysaffiliation{to}{Department of Computer Science, ETH Zurich, Switzerland}
\mlsysaffiliation{goo}{VMware Research and Department of Computer Science, UIUC, USA}

\mlsyscorrespondingauthor{Christoph Müller}{christoph.mueller@inf.ethz.ch}
\mlsyscorrespondingauthor{Francois Serre}{serref@inf.ethz.ch}

\mlsyskeywords{Machine Learning, Neural Network Verification, Formal Methods, Static Analysis}

\vskip 0.3in

\begin{abstract}
Certifying the robustness of neural networks against adversarial attacks is essential to their reliable adoption in safety-critical systems such as autonomous driving and medical diagnosis. Unfortunately, state-of-the-art verifiers either do not scale to bigger networks or are too imprecise to prove robustness, limiting their practical adoption.
In this work, we introduce \tool, a scalable verifier that can prove the robustness of significantly larger deep neural networks than previously possible. The key technical insight behind \tool is the design of custom, sound polyhedra algorithms for neural network verification on a GPU. Our algorithms leverage the available GPU parallelism and inherent sparsity of the underlying verification task.
\tool scales to large networks: for example, it can prove the robustness of a 1M neuron, 34-layer deep residual network in $\approx$ 34.5 ms. We believe \tool is a promising step towards practical verification of real-world neural networks.
\end{abstract}
]
\printAffiliationsAndNotice{\mlsysEqualContribution}

\section{Introduction}

\begin{figure}[ht]
    \begin{center}
    \includegraphics[scale=0.5]{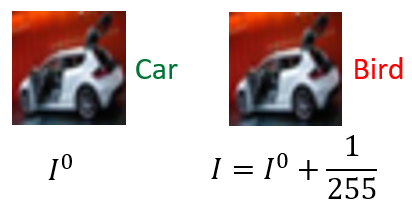}
    \end{center}
\vspace{-5mm}
    \caption{Image $I^0$ is classified correctly as a car by the neural network, while image $I$, obtained by increasing the intensity of each pixel in $I^0$ by $1/255$ is wrongly classified as a bird.}
    \label{Fi:intro}
\end{figure}

With the widespread adoption of deep neural networks in several real-world applications such as face recognition, autonomous driving, and medical diagnosis, it is critical to ensure that they behave reliably on a wide range of inputs. However, recent studies \cite{szegedy:13} have shown that deep networks are vulnerable to \textit{adversarial examples}, illustrated in \figref{intro}. Here, a neural network classifies an image $I^0$ correctly as a car. However, an adversary can increase the intensity of each pixel in $I^0$ by a small imperceptible amount to produce a new image $I$ that still looks like a car but the network incorrectly classifies it as a bird.

\textbf{Neural network robustness.}
Given this susceptibility to adversarial examples, recent years have seen increased interest in automated methods that can certify robustness of neural networks, that is, to prove that adversarial examples cannot occur within a specified adversarial region \cite{katz:17, Ehlers:17, wong18a, Gehr:18}. A typical example of an adversarial region would be the $L_\infty$ ball of radius $\epsilon \in \mathbb{R}^+$ around an image $I^0$ \citep{Carlini:17}. The goal of certification then is to prove that all images in this region are classified correctly by the network (i.e., to the same label as $I^0$). Note that the adversarial region
usually contains an exponential (in image size) number of images, which makes exhaustive enumeration infeasible. For example, image $I^0$ in \figref{intro} contains $3,072$ pixels. If we consider a radius of $\epsilon=1/255$ around $I^0$, then the number of images in the adversarial set $L_{\infty}(I^0,\epsilon)$ is $3^{3072}$ (in our experiments we consider significantly larger $\epsilon$ values).

\textbf{Key challenge: scalable and precise verification.}
Because concrete enumeration is infeasible, neural network verifiers compute the output for all inputs in the adversarial region symbolically. These verifiers can be broadly classified as either exact or inexact. Exact verifiers typically employ mixed-integer linear programming (MILP) \cite{tjeng:18}, SMT solvers \cite{katz:17,Ehlers:17,bunel:18,Marabou:19} and Lipschitz optimization \cite{ruanhk:18}. They are computationally expensive and do not scale to the network sizes considered in our work.
To address this scalability issue, inexact verifiers compute an over-approximation of the network output. Due to this approximation, a verifier may fail to prove the network robust when it actually is. Inexact verifiers are typically based on abstract interpretation \cite{Gehr:18,mirman:18, Singh:18,Singh:19}, duality \cite{Dvijotham:18,wong18a}, linear approximations \cite{Weng:18, zhang:18, cnncert:19, salman:19, zhang2020towards, Reluval:2018,IBP:18, Lirpa:20, TranBXJ:20}, and semi definite relaxations \cite{aditi:18,DathathriDKRUBS:20}. There are also methods \cite{shiqi:18,singhrobustness:19,singh2019krelu, Tjandraatmadja:20} that combine both exact and inexact approaches aiming to be more scalable than exact methods while improving the precision of inexact methods.

There is a trade off between scalability and the degree of over-approximation of inexact verifiers. More precise, inexact verifiers \cite{wong18a,Gehr:18,Singh:18,Singh:19,Weng:18, zhang:18, cnncert:19, salman:19, aditi:18,shiqi:18,singhrobustness:19, TranBXJ:20} scale to medium-sized networks ($\approx 100$K neurons) or verify weaker robustness properties (e.g. brightness \citep{Pei:17}) but cannot handle the networks and properties (e.g. $L_{\infty}$-norm) that our work can ($\approx 1$M neurons). On the other hand, more approximate verifiers \cite{mirman:18,Reluval:2018,IBP:18, zhang2020towards, Lirpa:20} scale to bigger networks but lose too much precision and fail to prove robustness, which limits their applicability. Thus, a key challenge is to design neural network verifiers that scale to large networks yet maintain the precision necessary to certify meaningful robustness guarantees.

\textbf{Scalable, precise and sound verification on a GPU.}
In this work, we present \tool, a new neural network verifier that addresses the above challenge via algorithms that leverage the processing power of GPUs. Concretely, \tool: (i) introduces a method that enables the fine-grain data parallelism needed to benefit from GPUs, (ii) is memory efficient and can fit into GPU memory (which is much smaller than that of a CPU), and (iii) is sound for floating point arithmetic, capturing all results possible under different rounding modes and orders of execution of floating point operations, thus handling associativity correctly (important concern, as recent verifiers which are unsound for floating-point have been shown vulnerable to such attacks \cite{jia2020exploiting, zombori:21}).

\tool is based on the state-of-the-art \deeppoly relaxation \cite{Singh:19} equipped with new, custom algorithms which exploit the underlying sparsity, and use a novel stopping criteria that can decrease runtime without compromising accuracy.

We note that it is possible to implement \deeppoly on a GPU using off-the-shelf libraries such as PyTorch \cite{pytorch:17} and Tensorflow \cite{tensorflow:15} as in \cite{zhang2020towards, Lirpa:20}. Unfortunately, these frameworks cannot exploit the sparsity patterns produced by \deeppoly, resulting in implementations that lack the performance and memory efficiency needed for handling the large networks considered in our work. 

\textit{Main contributions.} Our main contributions are:
\begin{itemize}
    \item New algorithms to efficiently parallelize the state-of-the-art \deeppoly relaxation on GPUs, enabling fast and precise verification of networks with up to $\approx 1$M neurons.
    \item A complete floating-point-sound CUDA implementation in a verifier called \tool that handles fully-connected, convolutional, and residual networks. Our code is available as part of the ERAN framework at \url{https://github.com/eth-sri/eran}.
    \item An experimental evaluation of \tool demonstrating its effectiveness in proving the robustness of neural networks beyond the reach of prior work.
\end{itemize}

We note that while we use \tool for proving robustness against intensity perturbations in this work, \tool is more general and can be used to certify other properties including safety \cite{katz:17}, fairness \cite{RuossBFV:20}, and robustness against geometric \cite{BalunovicBSGV:19,Spatial:21}, contextual \cite{paterson:21}, and generative \cite{mirman:20} perturbations.

\section{Background and Notation} \label{Se:background}

We now introduce the necessary background on both neural network robustness and the \deeppoly relaxation.

\mypar{Classification network} For simplicity, the networks we consider here are built from a composition of two kinds of layers: the \emph{affine} and the \emph{ReLU} layer. We use the word \emph{neuron} for the abstract node in such a layer, and we denote with $x_i^\layerno$ the $i^{\text{th}}$ neuron in the $\layerno^\text{th}$ layer $x^\layerno$.
The affine layers such as fully-connected, convolutional, and residual layers perform an affine mapping $x^{\layerno}=A\cdot x^{\layerno-1} + b,$ where $A=(a_{i,j})$ is a matrix and $b=(b_i)$ is a vector. The ReLU layer trims negative values element wise: $x^{\layerno}_i=\max(x^{\layerno-1}_i,0).$
A given input image $I$ is assigned to the input layer $x^0$, and evaluated successively through the different layers. The neuron index in the final layer with the highest value yields the inferred category.

\mypar{\boldmath$L_\infty$-norm based robustness properties}
Given an image $I^0$ correctly classified by the network, and a number $\epsilon > 0$, the \emph{adversarial region} $L_\infty(I^0,\epsilon)$ is the set of images $I$ for which each pixel $i$ differs by at most $\epsilon$ from the corresponding one in $I_0$: $||I_i-I^0_i||_\infty\leq\epsilon.$ The objective of a verifier is to prove that all images in this region classify correctly. 

\mypar{\deeppoly analysis}
The \deeppoly  \cite{Singh:19} relaxation associates four bounds with every neuron $x_i^{\layerno}$: (i) lower and upper polyhedral bounds of the form $\sum_{j}a_{i,j}\cdot x_j^{k}+c \leq x_i^{\layerno}$ and $x_i^{\layerno}\leq \sum_j a'_{i,j}\cdot x_{j}^{k} +c'$, respectively, where $0\leq k<\layerno$, and (ii) interval bounds $l_i^{\layerno} \leq x_i^{\layerno} \leq u_i^{\layerno}$, where $a_{i,j},a'_{i,j},c,c',l_i^{\layerno},u_i^{\layerno} \in \mathbb{R}$. We refer to $\sum_{j}a_{i,j}\cdot x_j^{k}+c$ and $\sum_j a'_{i,j}\cdot x_j^{k} +c'$ as the lower and upper polyhedral expressions, respectively. The polyhedral bounds for each $x_i^{\layerno}$ are obtained by modeling the effects of affine transformation and ReLU as follows:
\begin{itemize}
\item{The affine transformation $x_i^{\layerno}=\sum_j a_{i,j}^{\layerno-1} \cdot x_j^{\layerno-1} + b_i$ adds the bounds  $\sum_j a_{i,j}^{\layerno-1} \cdot x_i^{\layerno-1}  + b_i \leq x_i^{\layerno} \leq \sum_j a_{i,j}^{\layerno-1}\cdot x_i^{\layerno-1} + b_i$. Thus \deeppoly handles affine layers exactly.}
\item{$x_i^{\layerno}= \text{ReLU}(x_i^{\layerno-1})$ adds the bounds $\alpha^{\layerno}_i \cdot x_i^{\layerno-1}+ \beta_i^{\layerno} \leq x_i^{\layerno} \leq \gamma_i^{\layerno} \cdot x_i^{\layerno-1} + \delta_i^{\layerno}$ where the constants $\alpha^{\layerno}_i, \beta^{\layerno}_i, \gamma^{\layerno}_i, \delta^{\layerno}_i$ are determined by the concrete bounds $l_i^{\layerno-1} \leq x_i^{\layerno-1} \leq u_i^{\layerno-1}$.
If $l_i^{\layerno-1} > 0$ or $u_i^{\layerno-1} \leq 0$, then the \deeppoly analysis is exact, otherwise it over-approximates.
}
\end{itemize}

The tightness of the concrete bounds of the neurons from the affine layers that are input to ReLU layers affects the precision of the \deeppoly analysis as the bounds determine whether the ReLU is handled exactly and otherwise affect the level of imprecision in case its effect is over approximated. \deeppoly computes tight concrete bounds for each $x_i^{\layerno}$ in an affine layer by maximizing and minimizing its value with respect to the set of polyhedra and interval bounds over the neurons in all previous layers already computed by \deeppoly. \deeppoly solves both these linear programs approximately (for scalability reasons) using a greedy algorithm called \textit{backsubstitution} (not to be confused with back-propagation) which is the main bottleneck of the analysis and therefore the focus of our work.


\mypar{Bottleneck Backsubstitution}
The backsubstitution algorithm for computing an upper bound $u_i^{\layerno}$ (the lower bound is computed analogously) for neuron $x_i^{\layerno}$ in an affine layer starts with the upper polyhedral bound $x_i^{\layerno} \leq  \sum_j a_{i,j}^{\layerno-1}\cdot x^{\layerno-1} + b_i$ added by the affine transformation. It then substitutes the concrete upper or lower bounds for each $x^{\layerno-1}$ depending on the sign of the coefficient $a_{i,j}$ obtaining a candidate upper bound.
Next, it substitutes for each $x^{\layerno-1}$ the corresponding upper or lower polyhedral bound (again depending on the sign of $a_{i,j}$) defined over the neurons in layer $\layerno-2$. This yields a new polyhedral bound for $x_i^{\layerno}$ now defined over the neurons in layer $\layerno-2$. It again uses concrete bounds for the neurons in $x^{\layerno-2}$ to compute another candidate bound.

The algorithm repeats this step until it reaches the input layer. The result is the smallest candidate among the bounds computed at each step.
Since backsubstitution only involves reading data from the previous layers, it can be executed in parallel for all neurons in a given layer $\layerno$, which is ideal for GPU parallelization. For simplicity, we will focus on the most expensive step of backsubstitution for the remainder of this paper: computing new polyhedral bounds for $x_i^{\layerno}$ when the polyhedral bounds for the neurons $x_j^k$ with $k < \layerno$ to be substituted are the result of an affine transformation in fully-connected, convolutional, or residual layers. Without loss of generality, we therefore ignore the ReLU layers and consider neural networks as just a sequence of affine layers.


\begin{figure}[t]
\begin{center}
\includegraphics[scale=0.25]{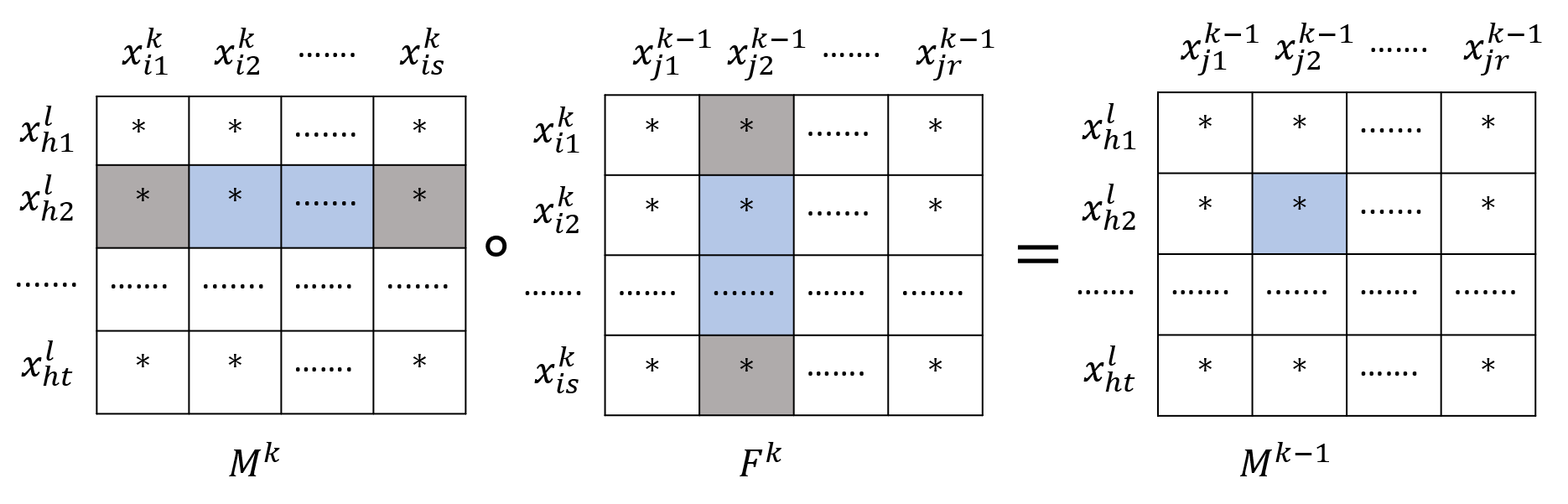}
\end{center}
\vspace{-3mm}
 \caption{Backsubstitution in polyhedra bounds for $\layerno$-th layer neurons. We omit the constant term in the bounds for simplicity.}
    \label{Fi:mm}
\vspace{-2mm}
\end{figure}

\mypar{Backsubstitution as matrix multiplication}
The left matrix $M^k$ in \figref{mm} encodes bounds for $\layerno$-th layer neurons with polyhedral expressions defined over the neurons in layer $1\leq k<\layerno$. The center matrix $F^k$ encodes constraints for $k$-th layer neurons defined over the neurons in layer $k-1$. We focus on the computation of the entry $(h2,j2)$ (shown in blue) in the result matrix $M^{k-1}$. The corresponding entries of $M^k$ and $F^k$ used for computing $(h2,j2)$ are subset of their $h2$-th row and $j2$-th column respectively (also shown in blue). The entry $(h2,j2)$ encodes the coefficient for neuron $j2$ of layer $k-1$ in the constraint for $x_{h2}^\layerno$. The substitution (as defined above) computes $(h2,j2)$ by multiplying blue each entry $(h2,i)$ in $M^k$ with the blue entry $(i,j2)$ of $F^k$ where $1 \leq i \leq s$. Each multiplication result represents a term involving $x_{j2}^{k-1}$ obtained by substituting the expression for $x_i^k$ in the constraint for $x_{h2}^\layerno$. The results are then summed which causes cancellation. This computation can be seen as multiplying the $h2$-th row of $M^k$ with the $j2$-th column of $F^k$ and the overall computation thus is a matrix multiplication $M^{k-1} = M^k \cdot F^{k}$.

We note that for fully-connected layers, all entries in the $h2$-th row and $j2$-th column are needed for computing $(h2,j2)$, while the convolutional and residual layers require a smaller subset. Identifying this subset is key to achieving the compute and memory efficiency required for obtaining a precise and scalable analysis.
We design custom algorithms tailored to exploit the sparsity patterns observed when handling convolutional and residual layers. We note that while matrix multiplication can be easily parallelized on GPUs, the standard algorithms \cite{zhang2020towards, Lirpa:20} are not memory and compute efficient for our task and run out of memory on medium-sized benchmarks ($\approx 170K$ neurons). Further, to ensure floating point soundness we perform all computations in interval arithmetic, which prevents the use of existing libraries. 


\textbf{Asymptotic cost.}
Consider a neural network with $n$ affine layers and with each layer containing at most $N$ neurons. The backsubstitution tasks for all neurons at an intermediate layer $\layerno \leq n$ perform a matrix multiplication in $\mathcal{O}(N^3)$ for all preceding affine layers (the ReLU layers have quadratic cost) resulting in an overall cost of $\mathcal{O}(\layerno \cdot N^3)$. Because backsubstitution is performed for every layer of the network, the \deeppoly algorithm requires $\mathcal{O}(n^2 \cdot N^3)$ operations.

\section{Robustness Verification on GPUs: Concepts and Algorithms} \label{Se:overview}

In this section, we introduce two key concepts that we exploit to design and implement an efficient DeepPoly-based GPU algorithm for verifying deep neural networks. The notion of \emph{dependence set} allows us to harness the sparsity of convolutional and residual layers to speedup backsubstitutions, while an \emph{early termination criterion} allows us to skip computations that would not improve results.


\subsection{Dependence set}
\begin{figure}
    \begin{center}
    \includegraphics[scale=0.24]{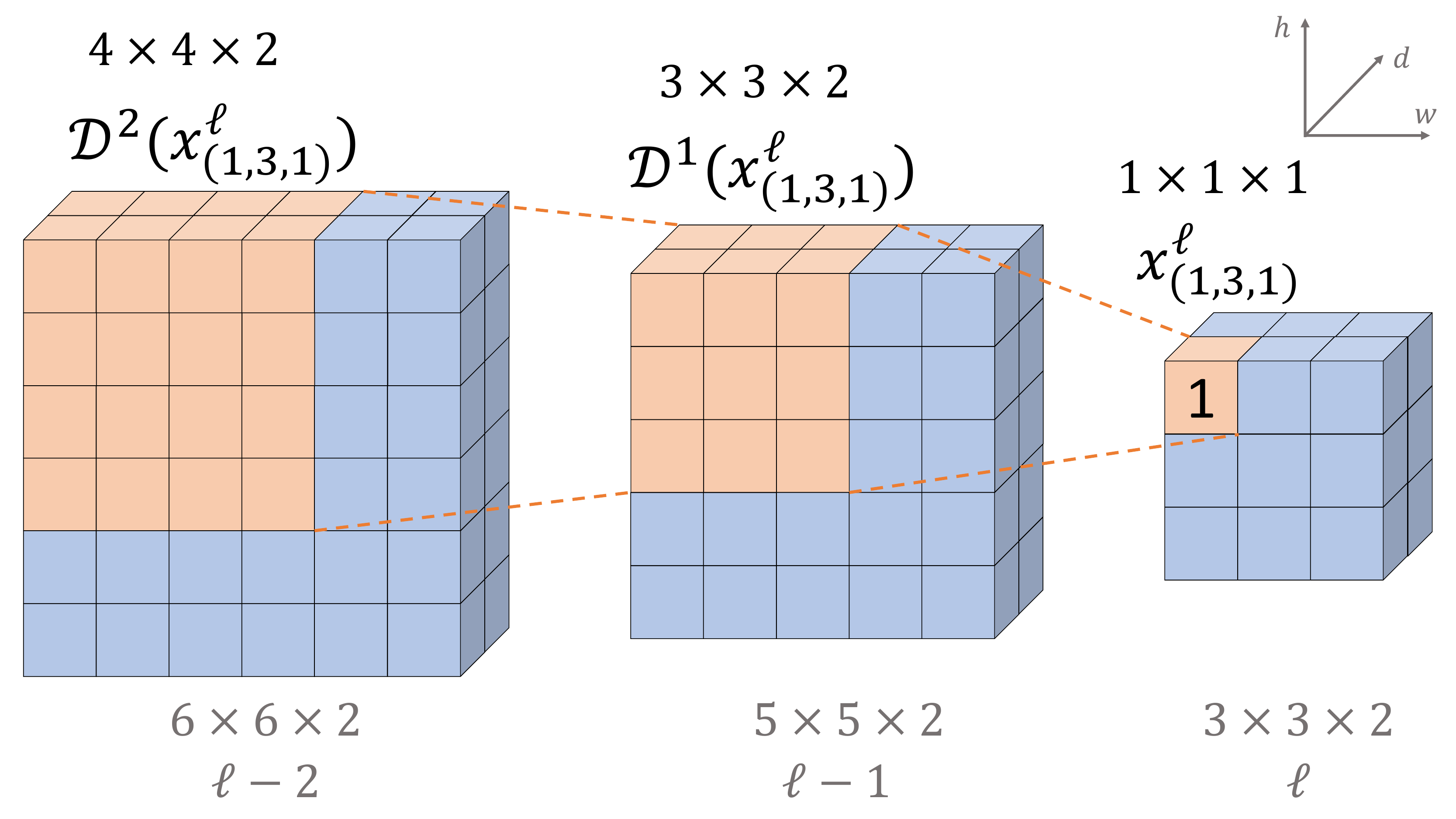}
    \end{center}
    \caption{Backsubstitution from a single neuron in layer $\layerno$ to layers $\layerno - 1$ and $\layerno - 2$. The number of neurons in layers $\layerno$, $\layerno-1$, $\layerno-2$ are $3\times3\times2$, $5\times5\times2$, and $6\times6\times2$ respectively.}
    \label{Fi:motivation}
\end{figure}

The core concept for exploiting sparsity in convolutional layers in our algorithms is the dependence set. Before defining it formally, we illustrate it on an example of backsubstitution through two convolutional layers (backsubstitution through fully-connected layers can be implemented as a dense matrix-matrix multiplication as explained in \secref{background}).
%
 We denote the neuron $i$ in a convolutional layer $\layerno$ as $x_i^{\layerno}= x_{w,h,d}^{\layerno}$, where $w,h,d$ are its indices in the width, height, and depth dimensions, respectively.
The rows of the matrix $M^k$ ($1 \leq k < \layerno$) depicted in \figref{mm} are the intermediate results of $ht$ many independent backsubstitutions, one for each neuron in layer $\layerno$. We show one such single-neuron backsubstitution in \figref{motivation} for neuron $x_{1,3,1}^\layerno$. In our example, layer $\layerno$ has size $3\times 3\times2$, whereas the previous layer $\layerno - 1$ has size $5\times5\times2$. The convolution operation with filters having $w$ and $h$ dimension $3 \times 3$, creates constraints for the neuron in layer $\layerno$ with a subset of the neurons in layer $\layerno-1$ that are part of a $3 \times 3 \times 2$ block, as shown in layer $\ell - 1$ in \figref{motivation}.

We call this set of neurons in layer $\layerno - 1$ the first dependence set of $x_{(1, 3, 1)}^\layerno$. Note that the first dependence set of $x_{(1, 3, 1)}^\layerno$ and $x_{(1, 3, 2)}^\layerno$ is the same. The second dependence set of $x_{(1, 3, 1)}^\layerno$, also shown in \figref{motivation}, has size $4 \times 4 \times 2$ (filters between layer $\layerno - 2$ and $\layerno - 1$ have $w$- and $h$-dimension $2 \times 2$). The second dependence set of $x_{(1, 3, 1)}^\layerno$ is obtained by taking the neurons in the output of the first dependence set and then for each neuron in this output, adding its corresponding first dependence set to the final output. 

The dependence sets identify the dense submatrices (blue entries in $M^k$ and $F^k$ in \figref{mm}) needed for computing backsubstitution through convolutional and residual layers. This enables a compute and memory-efficient GPU implementation that leverages dense matrix-matrix operations for high performance gains.
Next we define dependence sets formally and then present our algorithms.

\mypar{Network DAG}
We first define the \emph{network DAG} associated with a neural network.
In a network DAG $(\mathcal{V},\mathcal{E})$, $\mathcal{V}$ is the set of all neurons. Two neurons are connected by a directed edge $(x_j^k,x_i^\layerno)\in\mathcal{E}$ if $x_j^k$ is directly needed to compute $x_i^\layerno$.
More formally, $(x_j^{k},x_i^\layerno)\in\mathcal{E}$ if layer $k$ is an immediate predecessor (contains inputs) of layer $\layerno$ and
\begin{itemize}
    \item $\layerno$ is a convolutional layer and $x_j^{k}$ is in the window for computing $x_i^{\layerno}$, or
    \item $\layerno$ is a ReLU or a residual layer and $j=i$, or
   	\item layer $\layerno$ is fully-connected.
\end{itemize}
Note that for fully-connected and convolutional architectures, we have $k=\layerno-1,$ while for a residual network, layer $\layerno$ can have multiple immediate predecessors $k<\layerno$.
\mypar{Formal definition} \label{Se:dependence_set}
The \emph{first dependence set} of a neuron $x_i^\layerno$ collects all its immediate predecessors in the network DAG:
\begin{equation}
    \dependence{1}(x_i^\layerno) = \{ x_j^{k}~|~(x_j^k, x_i^{\layerno}) \in \mathcal{E} \},
\end{equation}
Similarly for a set of neurons $\mathcal{X}^\layerno$ in the same layer $\layerno$:
%


\begin{equation}
    \dependence{1}(X^\layerno) = \bigcup_{x_i^\layerno \in X^\layerno} \dependence{1}(x_i^\layerno)
\end{equation}

We extend this concept recursively. The \emph{$m$-th dependence set}, $m\geq 2$, of $x_i^\layerno$ is the first dependence set of $\dependence{m-1}(x_i^\layerno)$:
\begin{align}
    \dependence{m} (x_i^\layerno) = \dependence{1}(\dependence{m-1}(x_i^\layerno)) 
\end{align}
and the definition of $\dependence{m} (\mathcal{X}^\layerno)$ is analogous. We also define the \emph{zeroth dependence set} as $\dependence{0}(x_i^\layerno)=\{x_i^\layerno\}$.


%
%

During \deeppoly analysis, all neurons appearing in the polyhedral bounds obtained when backsubstituting iteratively on the polyhedral constraints for $x_i^\layerno$ are available in the different sets $\dependence{\layerno-k}(x_i^\layerno)$ with $k=\layerno-1,\hdots,0$. The expression in the initial bound contains neurons from $\dependence{1}(x_i^\layerno)$ and we call it step $1$ of backsubstitution. Step $2$ substitutes for each neuron in $\dependence{1}(x_i^\layerno)$, the polyhedral bound defined over the neurons in $\dependence{2}(x_i^\layerno)$ resulting in a new bound for $x_i^\layerno$ defined over the neurons in  $\dependence{2}(x_i^\layerno)$.
Continuing analogously, we see that $\dependence{\layerno-k}(x_i^\layerno)$ contains the neurons appearing in the bounds after $\layerno-k$ steps.
In \secref{implementation}, we exploit the structure of the convolutional layers to derive recursive expressions for computing $\dependence{\layerno-k}(x_i^\layerno)$ that enable fast computation with negligible overhead. Next, we discuss the backsubstitution for the different network types in greater detail.





\mypar{Efficient backsubstitution for convolutional networks}\label{Se:ov_cnn}
Naively using a dense matrix-matrix multiplication for a backsubstitution starting at a convolutional layer $\layerno$ is very memory and compute inefficient.
First, the majority of computations are not needed since the filters in the convolutional layers are sparse and thus the filter matrix $F^k$ of \figref{mm} consists of mostly zeroes. Additionally, it is not memory efficient since many coefficients in matrices $M^k$ and $M^{k-1}$ of \figref{mm} will be zero during backsubstitution.

\textbf{Key idea.} The neurons in the polyhedral expression for $x_i^\layerno$ after $\layerno -k$ backsubstitution steps ($0\leq k < l$) are in the dependence set $\dependence{\layerno-k}(x_i^\layerno)$. For an efficient implementation on GPUs, utilizing the dependence set, we can flatten the needed coefficients into a dense matrix to perform the backsubstitution again efficiently as matrix-matrix multiplication.
This will be detailed in Section~\ref{Se:implementation}.

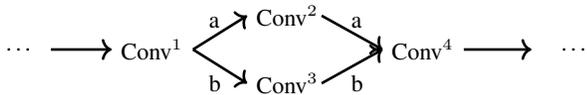
\begin{figure}[h!t]
\centering
\begin{tikzpicture}[scale=0.9]
\tikzstyle{style2}[black]=[draw=#1,fill=none,opacity=1.0,line width=1,->]

\begin{scope}
\coordinate (p1) at (0.3, -2) ;
\coordinate (p2) at (2, -2) ;
\coordinate (p3) at (4, -1.5) ;
\coordinate (p4) at (4, -2.5) ;
\coordinate (p5) at (6, -2) ;
\coordinate (p6) at (8.5, -2) ;
\coordinate (p7) at (3.3, -1.6) ;
\coordinate (p8) at (3.3, -2.5) ;
\coordinate (p9) at (5.4, -1.6) ;
\coordinate (p10) at (5.4, -2.5) ;

\filldraw[style2] (1.1,-2) -- (2,-2);
\filldraw[style2] (3.2,-2) -- (4,-1.5);
\filldraw[style2] (5.1,-1.5) -- (6,-2);
\filldraw[style2] (3.2,-2) -- (4,-2.5);
\filldraw[style2] (5.1,-2.5) -- (6,-2);
\filldraw[style2] (7.2,-2) -- (8.2,-2);

		\fill[black,font=\small]
            (p1) node [right] {\dots}
            (p2) node [right] {Conv$^1$}
            (p3) node [right] {Conv$^2$}
            (p4) node [right] {Conv$^3$}
            (p5) node [right] {Conv$^4$}
            (p6) node [right] {\dots}
            (p7) node [right] {a}
            (p8) node [right] {b}
            (p9) node [right] {a}
            (p10) node [right] {b};
\end{scope}

\end{tikzpicture}

    \caption{Simplified residual architecture without ReLU layers.}
\label{Fi:residual_layer}
\end{figure}

\mypar{Dependence set and residual networks}\label{Se:ov_res}
To simplify the exposition of our ideas and without loss of generality, we assume that the width of the residual network is two, i.e., a layer has no more than two immediate predecessors or successors. An example of such an architecture is in \figref{residual_layer} which shows a residual block consisting of one convolutional layer in each branch with all ReLU layers removed for simplicity.

For simplicity, we assume that the two branches of a residual block have the same length and call them $a$ and $b$. In \figref{residual_layer} branches $a$ and $b$ contain the Conv$^2$ and Conv$^3$ layer, respectively.
Naturally, the layer at the head of the residual block (Conv$^1$ in  \figref{residual_layer}) has two successors while the one at exit (Conv$^4$ in \figref{residual_layer}) has two predecessors.

The first dependence set of a neuron $x_i^\layerno$ in a layer at the exit of a residual block (e.g., Conv$^4$ in \figref{residual_layer}) contains neurons from both branches (subsets of layers Conv$^2$ and Conv$^3$ in \figref{residual_layer}). The resulting dependence set can be written as:
\begin{equation}
    \dependence{1}(x_i^\layerno) = \dependence{(1, a)}(x_i^\layerno) \cup \dependence{(1, b)}(x_i^\layerno),
\end{equation}
where $\dependence{(1, a)}(x_i^\layerno)$ and $\dependence{(1, b)}(x_i^\layerno)$ are the first dependence sets of $x_i^\layerno$ with respect to branches $a$ and $b$, respectively.

In our algorithm, we leverage the above partition of the first dependence set to backsubstitute through both branches independently and then join the independent backsubstitutions at the head of the residual block (in our case Conv$^1$) by adding the coefficients of the expressions neuron by neuron. For this, the two resulting dependence sets coming from the two residual branches, which do not necessarily have the same size, need to be overlapped correctly. We omit these details due to lack of space.

\subsection{Early termination} The \deeppoly backsubstitution algorithm can be terminated early if the following criterion is met.

\mypar{Termination criterion} The polyhedral approximation of a ReLU layer is exact if $0$ is not strictly included within its bounds. In this case, no additional precision can be gained for this neuron by backsubstituting further.

An efficient implementation of DeepPoly should therefore filter those neurons out of the backsubstitution that satisfy this termination criterion. With the formalism introduced in Section~\ref{Se:background}, this amounts to removing a selection of rows out of the matrix $M^k$. We will propose a method to perform this operation efficiently with a shared memory machine model in Section~\ref{Se:implementation}.

\section{\tool} \label{Se:implementation}
We now explain our \tool algorithm. We first explain how to maintain floating point soundness using interval arithmetic. Next, we discuss the implementation of the early termination criterion.  Then we discuss how to compute the size and elements of the dependence set $\dependence{\layerno - k}(x_i^\layerno)$ of $x_i^\layerno$ in layer $k < \layerno$ for convolutional layers. We use this set in our parallel algorithm for the backsubstitution.


\subsection{Floating point soundness} \label{Se:soundness}

An essential property and major challenge is to ensure that our certification guarantees are valid under floating-point arithmetic \cite{jia2020exploiting,zombori:21} where round off errors are frequent and common mathematical properties such as associativity do not hold. To be floating point sound, our analysis output should contain all results possible under different rounding modes and execution orders of operations. To achieve this, we replace the scalar coefficients of our polyhedra bounds with intervals. Therefore, our bounds actually describe a set of polyhedra instead of a single polyhedron.

To ensure soundness under all rounding modes, all floating point operations on intervals are performed such that the lower bound is always rounded towards $-\infty$ and the upper bound towards $+\infty$. This particularly prevents the use of standard BLAS libraries. Our matrix-matrix multiplication procedure is built around a custom multiply-add operation, and uses the cutlass template library for tiling.
In addition, \tool takes into account the error that may occur during inference, because of the lack of associativity for floating point operations. This is done by systematically taking the next representable floating value for the terms of all summations, in the direction that will over-approximate the error as described in detail in \cite{rounding}. Overall, ensuring floating point soundness doubles the memory requirement and more than doubles the number of floating point operations needed.



\subsection{Implementing early termination} \label{Se:early_termination}
The original DeepPoly algorithm performs a complete backsubstitution for all the input neurons of ReLU layers. In this part, we describe the changes we introduced in \tool to fully exploit the early termination criterion described in Section~\ref{Se:overview}.

In order to have a first approximation of the interval bounds of hidden neurons, a forward interval analysis is performed as a preliminary step. Then, a regular DeepPoly analysis is performed, with the particularity that when a ReLU layer is encountered, its inputs (rows in the matrix $M^k$) are not directly backsubstituted. Instead, an intermediate matrix $M'^k$ containing only the neurons that do not match the termination criterion is created, along with an array containing the indices of the corresponding rows in the original matrix $M^k$ (we will explain this step later). Then, a backsubstitution is performed on the matrix $M'^k$, and the resulting interval bounds are assigned to their corresponding neurons, using the array. Finally, a forward interval analysis updates the approximations of the following layers, before regular DeepPoly analysis resumes.

By construction, \tool visits ReLU layers in a topological order with respect to the network DAG. This ensures that all backsubstitutions only use the best possible polyhedral approximation of their ancestors, thus guaranteeing the same result as the original algorithm. In addition, during these backsubstitutions, concrete bounds are re-evaluated regularly, and the rows of the neurons that match the termination criterion are removed from $M'^k$.

In the worst case, \tool computes all backsubstitutions completely, and has similar performance as if this optimization was not implemented, as the additional steps have a negligible runtime with respect to backsubstitutions. However, in many practical cases (as in \secref{experiments}), significantly fewer backsubstitutions are actually computed, yielding a significant speedup.

\mypar{Removing rows from a matrix in a shared memory context}
To create $M'^k$ and the corresponding index array, each thread of the GPU is associated with one row of $M^k$, and checks whether the termination criterion is met for that row. A parallel prefix sum is then performed between all threads, with the value $0$ if the termination is met, and $1$ otherwise. This way, each thread associated with a non-terminated neuron receives a unique integer $i$ ranging between $0$ and the number of non-terminated neurons. Finally, each thread associated with a non-terminated neuron copies its corresponding row at the $i^\text{th}$ row of $M'^k$, and writes its index at the $i^\text{th}$ place of the array.

\mypar{Memory management}
For larger networks, the matrix $M^k$ may not entirely fit in GPU memory. In these situations, the intermediate matrix $M'^k$ can be used to sequentially backsubstitute chunks of $M^k$ that are small enough to fit in memory.


\subsection{Dependence sets for convolutional networks}

To simplify the exposition and without loss of generality, we assume that all parameters of convolutional layers have the same value in horizontal $h$ and vertical $w$ directions, such as filter sizes $f_w^k = f_h^k = f^k$, strides $s_w^k = s_h^k = s^k$ and the padding $p_w^k = p_h^k = 0$.

In \figref{motivation} we have seen examples of the first and second dependence set of a neuron in a convolutional layer. Now we derive the general equations for the size and offset of the elements of $(\layerno - k)$-th dependence set $\dependence{\layerno - k}(x_i^\layerno)$ as a subset of the neurons in a convolutional layer $0 \leq k<\layerno$.
$\dependence{\layerno - k}(x_i^\layerno)$ is a cuboid and we compute the size of the set $\dependence{\layerno - k}(x_i^\layerno)$ along the height, width and depth direction separately.
We note that the $k$-th dependence set, given $k>0$, is always dense in the depth direction for convolutional layers, so the size of $\dependence{\layerno - k}(x_i^\layerno)$ in the depth dimension is equal to the number of channels of layer $k$. Because of our symmetry assumption for the $w$ and $h$ directions of convolutional parameters, the width and the height of $\dependence{\layerno - k}(x_i^\layerno)$ are equal, and we denote it with $W^{\layerno - k}$. The following recurrence computes $W^{\layerno - k + 1}$ given $W^{\layerno - k}$:
\begin{eqnarray}
W^0 & = & 1,\nonumber \\
W^{\layerno - k + 1} & = & (W^{\layerno - k} - 1) \cdot s^k + f^k, k = \layerno\dots 1.\label{eq:size}
\end{eqnarray}
For example, in \figref{motivation} we obtain $W^1=(W^0 - 1) \cdot 1 + 3 = 3$ for the first dependence set and $W^2= (W^1 - 1) \cdot 1 + 2 = 4$ for the second dependence set. The overall size of $\dependence{\layerno-k}$ is:
\begin{equation}
    |\dependence{\layerno - k}(x_i^\layerno)| = W^{\layerno - k} \cdot W^{\layerno - k} \cdot C^k,  \quad k = \layerno-1\dots 0.\label{eq:dependence_size}
\end{equation}
where $C^k$ is the number of channels of layer $k$.
We compute the neuron indices next.

The indices depend on the location of $x_i^\layerno$ in layer $\layerno$.  We only need to derive the position in the width and the height direction as all the corresponding channels of layer $k$ are in $\dependence{\layerno - k}(x_i^\layerno)$. Let the position of $x_i^\layerno$ in layer $\layerno$ be $i=(w^\layerno, h^\layerno, d^\layerno)$. Then the $w$- and $h$-positions of the neuron with the smallest coordinates in $\dependence{\layerno - k}(x_i^\layerno)$ are:
\begin{align}
    w^{\layerno - k} &= S^{\layerno - k} \cdot w^\layerno,   \label{eq:position_w} \\
    h^{\layerno - k} &= S^{\layerno - k} \cdot h^\layerno, \quad k = \layerno-1\dots 0.\label{eq:position_h}
\end{align}
where we introduced the quantity $S^{\layerno-k}$, which we call \emph{accumulated stride} computed via the following recurrence:
\begin{eqnarray}
S^0 & = & 1,\\
S^{\layerno - k + 1} & = & s^k \cdot S^{\layerno - k},\quad k = \layerno\dots 1. \label{eq:stride}
\end{eqnarray}
The extension to other padding modes is similar.
Using the the size and the position of $\dependence{\layerno - k}(x_i^\layerno)$ in layer $k$, we can now recursively compute, for $k = \layerno-1\dots 0$ the associated coefficients of the neurons in $\dependence{\layerno - k}(x_i^\layerno)$ occurring in the backsubstituted expression. We store these in a dense matrix called $M^{k}(x_i^\layerno)$. In each step these get modified by the backsubstitution:
\begin{flalign*}
M^{\ell - 1}(x_i^\layerno)  = & (a_1,a_2,\hdots,a_{|\dependence{1}(x_i^\layerno)|}),\\
M^{k - 1}(x_i^\layerno) =  &\texttt{GBC}(M^{k}(x_i^\layerno),\dependence{\layerno-k}(x_i^\layerno),\\
 &\dependence{\layerno-k+1}(x_i^\layerno),
F^k), \quad 1\leq k \leq \layerno-1.\label{eq:submatrix}
\end{flalign*}
$M^{\ell - 1}(x_i^\layerno)$ contains the coefficients corresponding to the neurons in the first dependence set in the initial polyhedra bound. We ignore the constant in the bound for simplifying our exposition. \texttt{GBC} (\tool Backsubstitution for Convolution) is our algorithm for handling a single step of a backsubstitution task in convolutional networks, shown in Algorithm \ref{alg:filter_sparse_affine} and explained below. $F^k$ is the bound matrix between the neurons in layer $k$ and $k-1$ generated during \deeppoly analysis (\figref{mm}).  As in \secref{background}, $F^k$ corresponds to the filter for convolutional layers. We next explain \texttt{GBC} in greater detail.


%
%
%
%
%

\subsection{Our algorithm for convolutional networks} \label{Se:main_algorithm}


To be memory and compute efficient on GPU, our algorithm should compute the backsubstitution of the bound matrix $M^k$ through one convolutional layer $k$ obtaining $M^{k-1}$ (as in \figref{mm}), but for each row of the $M^k$s, only iterate over the respective dependence sets $\dependence{\ell - k}_{hi}$ and $\dependence{\ell - k + 1}_{hi}$. Ideally it should be possible to implement the algorithm via a matrix-matrix multiplication. Algorithm \ref{alg:filter_sparse_affine} satisfies these requirements.

\begin{algorithm}
\caption{{GBC}($M^{\layerno-k}$, $\forall hi:\left(\dependence{\layerno-k}_{hi}, \dependence{\layerno-k+1}_{hi}\right)$,$F^k$)}
\label{alg:filter_sparse_affine}
\fontsize{9}{10}\selectfont
\begin{algorithmic}[1]
   \STATE $M^{k}, M^{k-1} \gets$ \text{coefficient matrices for layers $k$, $k-1$}
    \STATE $\dependence{\layerno-k}_{hi} \gets $ \text{$(\layerno - k)$-th dependence set of $x_{hi}^\layerno$}
    \STATE $(W^{\layerno - k}, W^{\layerno - k}, C^k) \gets$ \text{dimensions of $\dependence{\layerno-k}$}
    \STATE $\dependence{\layerno-k+1}_{hi} \gets $ \text{$(\layerno - k + 1)$-th dependence set of $x_{hi}^\layerno$}
    \STATE $(W^{\layerno - k + 1}, W^{\layerno - k + 1}, C^{k-1}) \gets$ \text{dimensions of $\dependence{\layerno-k+1}_{hi}$}
    \STATE $(f^k, f^k) \gets$ \text{filter size in $w$ and $h$ directions of layer $k$}
    \STATE $(s^k, s^k) \gets$ \text{strides in $w$ and $h$ directions for layer $k$}
    \STATE  $F^k\gets$ \text{4-D filter weight tensor of layer $k$ }
    \STATE $(f^k, f^k, C^k, C^{k-1}) \gets$ \text{dimensions of $F^k$}
    \color{myblue}\FOR{$hi \in (h1:ht)$} \color{black} \label{lst:line:outer_loop}
     \FOR{$(w,h) \in (0:W^{\layerno - k}, 0:W^{\layerno - k})$}\label{lst:wh_loop}
    \FOR{$(f, g) \in (0:f^k,0: f^k)$}\label{lst:fg_loop}
        \STATE $a = w \cdot s^k + f$
        \STATE $b = h \cdot s^k + g$
        \color{myblue}\FOR{$c \in (0 :C^{k-1})$}\label{lst:ck-1_loop} \color{black}
            \STATE $M^{\layerno - k + 1}_{hi}[a][b][c] = 0$
            \color{myorange}\FOR{$d \in (0:C^k)$}\label{lst:ck_loop} \color{black}
               \STATE $M^{k - 1}_{hi}[a][b][c] \pluseq M^k_{hi}[w][h][d] \cdot F^k[f][g][d][c]$\label{lst:update}
           \ENDFOR
         \ENDFOR
    \ENDFOR
  \ENDFOR
\ENDFOR
\end{algorithmic}
\end{algorithm}

\begin{table*}[t]
    \centering
    \caption{Neural networks used in our experiments.}
    \footnotesize
    \begin{tabular}{@{}l l llrrr@{}}
        \toprule
        Dataset  & Model & & Type  & \#Neurons & \#Layers & Training\\
        \midrule
        MNIST & \texttt{$6 \times 500$} & & Fully-connected & 3,010 & 6 & Normal\\
              &  \texttt{ConvBig} & & Convolutional & 48K & 6 & DiffAI\\
              &  \texttt{ConvSuper} & & Convolutional & 88K & 6 & Normal\\
             &  \texttt{IBP\_large\_0.2, IBP\_large\_0.4} & & Convolutional & 176K & 6 & \crown\\
        \midrule
        CIFAR10 &  \texttt{$6 \times 500$} & & Fully-connected & 3,010 & 6 & Normal\\
              &  \texttt{ConvBig} & & Convolutional & 62K & 6 & DiffAI\\
              &  \texttt{ConvLarge} &  & Convolutional & 230K & 6 & DiffAI\\
             &  \texttt{IBP\_large\_2\_255, IBP\_large\_8\_255} & & Convolutional & 230K & 6 & \crown\\
              &  \texttt{ResNetTiny} &  & Residual & 311K & 12 & PGD\\
              &  \texttt{ResNet18} &  & Residual  & 558K & 18 & PGD\\
              &  \texttt{ResNetTiny} &  & Residual & 311K & 12 & DiffAI\\
              &  \texttt{ResNet18} &  & Residual & 558K & 18 & DiffAI\\
              &  \texttt{SkipNet18} &  & Residual & 558K & 18 & DiffAI\\
              &  \texttt{ResNet34} &  & Residual & 967K & 34 & DiffAI\\
        \bottomrule
    \end{tabular}
    \label{Ta:networks}
\end{table*}

The outermost loop $hi \in (h1:ht)$ in line \color{myblue} 10 \color{black} iterates over the rows of $M^k$. In lines 11 and \color{myorange} 17 \color{black} the algorithm loops over the dimensions $(W^{\layerno - k}, W^{\layerno - k}, C^k)$ of $\dependence{\layerno-k}_{hi}$. Instead of iterating on the full range $(W^{\layerno - k + 1}, W^{\layerno - k + 1}, C^{k-1})$ of $\dependence{\layerno-k+1}_{hi}$ the sparsity of the convolution allows us to only loop over the dimensions $(f^k, f^k, C^{k-1})$ in lines 12 and \color{myblue} 15 \color{black}. This requires additional index computations in lines 13 and 14 for expressing addresses in layer $k-1$ in terms of those in layer $k$. Finally in line 18 all non-zero entries of $M^{k-1}$ are computed given $M^k$ and the filter $F^k$.

Next we discuss the parallelization strategy. A matrix-matrix multiplication always has one dimension which is collapsed ($(i1:is)$ dimension in \figref{mm}), and two dimensions which can be parallelized ($(h1:ht)$ and $(j1:jr)$ in \figref{mm}). One of these two parallel dimensions is consecutive in memory ($(j1:jr)$) while the other is not ($(h1:ht)$). We follow the same strategy for the convolutional case, where the dimension we collapse is the loop in line \color{myorange} 17 \color{black}. The dimensions to be parallelized are the loops in lines \color{myblue} 10 \color{black} and \color{myblue} 15 \color{black}, where the loop in line \color{myblue} 15 \color{black} is consecutive in memory, as can be seen in line 18 where $c$ is the inner dimension for the matrices $M^{k-1}$ and $F^k$.
All other loops are left serial.

Note that this algorithm can be understood as performing a separate transpose convolution for every $h_i$. This transpose convolution is a map from $\dependence{\layerno-k}_{hi}$ to $\dependence{\layerno-k + 1}_{hi}$.
To guarantee floating point soundness for our algorithm as discussed in \secref{soundness}, we optimize an interval-scalar matrix-matrix multiplication where the coefficients in $M^k$ are intervals and $F^k$ contains the scalar network weights.


\textbf{Algorithm for residual blocks.} Algorithm \ref{alg:filter_sparse_affine} can be used to backsubstitute through both branches of a residual block separately, as discussed in \secref{ov_res}. The resulting two coefficient matrices then need to be added element-wise. Again we omit the details for space reasons.

\textbf{Comparison to the parallel CPU implementation.}
Since the available parallelism of a modern CPU is at least an order of magnitude smaller compared to a GPU, the parallelized CPU implementation \cite{Singh:19} of \deeppoly processes fewer rows of $M^k$ than \tool in parallel. 
The CPU implementation exploits sparsity in the polyhedral expressions when performing backsubstitution from the convolutional layers by storing the polyhedral expressions with a sparse representation, storing neuron indices and the corresponding coefficient. This representation does not exploit the structure of the convolutional layers and is not suitable for SIMD parallelization. In contrast, we exploit structured sparsity in convolutional layers via dependence sets which allows us to create smaller dense submatrices that are suitable for SIMD parallelization.

\begin{table*}[h]
    \centering
    \caption{Experimental results for 10,000 images on fully-connected and convolutional neural networks: \crown vs.~\tool.}
    \footnotesize
    \begin{tabular}{@{}llrrrrrrrrr@{}}
        \toprule
        Dataset & Model & \#Neurons  &  $\epsilon$ & \#Candidates &  & \multicolumn{2}{c}{\#Verified} & & \multicolumn{2}{c}{Median runtime} \\
        \cmidrule{7-8} \cmidrule{10-11}
        & & & & &   &  \crown  & \tool&  & \crown  & \tool \\
        \midrule
        MNIST   &  \texttt{$6 \times 500$} & 3,010  & 8/255 & 9,844 & & 0 & 7,291 & &  130\mus & 9.06 ms\\
                &  \texttt{ConvBig} & 48K            &  3/10 & 9,703 & & 5,312 & 8,809 & & 220\mus & 537 \mus\\
                &  \texttt{ConvSuper}  & 88K       & 8/255 & 9,901 & & 0 & 8,885 & &  300\mus & 266 ms\\
                &  \texttt{IBP\_large\_0.2}     & 176K   & 0.258  & 9,895 & & 4,071 & 7,122 &  & 190\mus & 9.04 ms\\
                &  \texttt{IBP\_large\_0.4}     & 176K   &   3/10  & 9,820 & & 9,332 & 9,338 &  & 190\mus & 2.92 ms\\
        \midrule
        CIFAR10 &  \texttt{$6 \times 500 $}  & 3,010  & 1/500 & 5,607 & & 0 & 4,519 & & 200\mus & 8.04 ms\\
                &  \texttt{ConvBig}  & 62K          & 8/255 & 4,599 & & 1,654 & 2,650 & & 320\mus & 730 \mus\\
                &  \texttt{ConvLarge}  & 230K         & 8/255 & 4,615 & & 1,672 & 2,838 & & 900\mus & 4.54 ms\\
                &  \texttt{IBP\_large\_2\_255}  & 230K   &   2/255 & 7,082 & & 5,450 & 5,588 &  & 820\mus & 12.3 ms\\
                &  \texttt{IBP\_large\_8\_255}  & 230K   &   8/255 & 4,540 & & 3,289 & 3,298 &  & 270\mus & 3.83 ms\\
        \bottomrule
    \end{tabular}
    \label{Ta:result_small_crown}
\end{table*}

\textbf{Comparison to standard backpropagation.}
Backpropagation \cite{autodiff} is fundamentally different from \deeppoly backsubstitution because it computes a scalar loss function and propagates it back to update the network weights while backsubstitution propagates constraints backwards. Further, backpropagation is usually only performed starting from the last layer which typically contains fewer neurons than the intermediate layers. In contrast, \deeppoly's backsubstitution is performed starting from all layers in the network. Thus, we also have to backsubstitute starting from intermediate convolutional or residual layers which typically contain orders of magnitude more neurons than the last layer which makes balancing the compute and the memory efficiency of the backsubstitution on GPUs more challenging (\secref{overview}). Overall, based on the above factors, the \deeppoly backsubstitution is mathematically different, computationally more expensive, and more memory-demanding than backpropagation.

\begin{table}[t]
    \centering
    \caption{Experimental results for 500 images on fully-connected and convolutional networks: \deeppoly vs.~\tool.}
    \footnotesize
    \begin{tabular}{@{}lrrrr@{}}
        \toprule
        Model &  \#Cand. & \#Verif. & \multicolumn{2}{c}{Median runtime} \\
        \cmidrule{4-5}
        & &  & \deeppoly & \tool \\
        \midrule
        \texttt{$6 \times 500$}   & 493 & 334 & 8.3 s & 9.06 ms\\
        \texttt{ConvBig}          & 487 & 441 & 12 s & 537 \mus\\
        \texttt{ConvSuper}        & 495 & 428 &  271 s & 266 ms\\
        \midrule
        \texttt{$6 \times 500 $}  & 282 & 219 & 15 s & 8.04 ms\\
        \texttt{ConvBig}          & 226 & 127 & 38 s & 730 \mus\\
        \texttt{ConvLarge}        & 232 & 138 & 309 s & 4.54 ms\\
        \bottomrule
    \end{tabular}
    \label{Ta:result_small_deeppoly}
\end{table}

\begin{table*}[t]
    \centering
    \caption{Experimental results for 1,000 images on big CIFAR10 residual networks: our implementation of \crown vs.~\tool.}
    \footnotesize
    \begin{tabular}{@{}llll rrrrrrr@{}}
        \toprule
        Model & \#Neurons & Training & $\epsilon$ & \#Candidates & & \multicolumn{2}{c}{\#Verified}  & & \multicolumn{2}{c}{Median runtime} \\
        \cmidrule{7-8} \cmidrule{10-11}
         & & & & & & \crown & \tool & &\crown & \tool \\
        \midrule
          \texttt{ResNetTiny}           & 311K & PGD    & 1/500 & 768 & & 0    & 651 &  & 5.3 s &  11.7 s\\
                 \texttt{ResNet18}      & 558K & PGD    & 1/500 & 823 & & 0    & 648 &  & 60 s  &  397 s\\
        \midrule
                 \texttt{ResNetTiny}    & 311K & DiffAI & 8/255 & 371 & & 203  & 244 &  & 5 s   &  4.03 ms\\
                 \texttt{SkipNet18}     & 558K & DiffAI & 8/255 & 321 & & 114  & 260 &  & 40 s  &  16.5 ms\\
                 \texttt{ResNet18}      & 558K & DiffAI & 8/255 & 372 & & 138  & 268 &  & 26 s  &  16.9 ms\\
                 \texttt{ResNet34}      & 967K & DiffAI & 8/255 & 356 & & 126  & 229 &  & 59 s  &  34.5 ms\\
        \bottomrule
    \end{tabular}
    \label{Ta:result_big}
\end{table*}

\section{Experimental Evaluation} \label{Se:experiments}

We now demonstrate the effectiveness of \tool for the verification of big neural networks in terms of both precision (number of instances verified) and performance in terms of runtime. \tool is implemented in C++, supports 64-bit double and 32-bit single precision, and uses the CUDA library for GPU support and Cutlass for the template metaprogramming of matrix operations in CUDA. We compare the effectiveness of \tool against two state-of-the-art verifiers: the CPU parallelized version of \deeppoly \cite{Singh:19} and the GPU based CROWN-IBP (\crown) from \citep{zhang2020towards,Lirpa:20}. We note that \deeppoly has the same precision as \tool, however \tool is at least 190x faster, for some networks even 68'000x, than \deeppoly. \crown is implemented for GPUs and is more precise than interval bound propagation \citep{mirman:18, IBP:18} and more scalable than CROWN-FULL \citep{ zhang:18}. We note that CROWN-FULL also has the same precision as \deeppoly \citep{salman:19}, however its GPU implementation from \citep{zhang2020towards,Lirpa:20} runs out of memory on most of our networks, therefore we do not consider it. Thus \crown is the most precise existing verifier that can scale to the big neural networks used in our experiments.
We note that unlike \tool and \deeppoly, \crown is not floating point sound thus its verification results can be incorrect due to floating point errors \cite{jia2020exploiting, zombori:21}.

Our experimental results show that \tool improves over the state-of-the-art by providing the most precise and scalable verification results on all our benchmarks. We believe that the extra scalability and precision of \tool can also benefit state-of-the-art robust training methods \cite{Balunovic2020Adversarial,zhang2020towards} in the future as they depend on approximate verifiers for training.

\textbf{Neural networks.}
We used $16$ deep neural networks in our experiments as shown in \tabref{networks}. Out of these, $5$ are MNIST-based  \cite{Lecun:98} and $11$ are CIFAR10-based  \cite{Krizhevsky:09}. \tabref{networks} specifies the network architecture, the number of neurons, the number of layers and the training method for each network. There are $2$ fully-connected, $8$ convolutional and $6$ residual architectures in \tabref{networks}. The largest network in the table is \texttt{ResNet34} with $34$ layers and $\approx$1M neurons.

Regarding training, (i) $7$ of our networks were trained using DiffAI \cite{mirman:18,mirman:19} and $4$ with \crown \cite{zhang2020towards}, both of which perform \textit{provably robust adversarial training}, (ii) $2$ of our CIFAR10 networks were trained using Projected Gradient Descent (PGD) \cite{madry:17,dong:18}, which amounts to \textit{empirically robust adversarial training}, and (iii) the remaining $3$ networks were trained in a standard manner. Both methods, (i) and (ii), aim to increase the robustness of the resulting neural network which results in a loss of standard accuracy.

In the following we will refer to the non-residual networks as medium networks and to the residual networks as big networks.

\textbf{Experimental setup.}
All our experiments for \crown and \tool were performed on a 2.2 GHZ 10 core Intel Xeon Silver 4114 CPU with $512$GB of main memory. The GPU on this machine was an Nvidia Tesla V100 GPU with $16$GB of memory. The PyTorch version used for running \crown was 1.3.0 and the CUDA version for \tool was 11.0. The experiments for the (prior) CPU version of \deeppoly were performed on a faster 2.6 GHz 14 core Intel Xeon CPU E5-2690 with $512$GB of memory.

\textbf{Benchmarks.}
For fully-connected and convolutional networks, we consider the full MNIST and CIFAR10 test sets. For the bigger residual networks, we selected the first $1,000$ images from the respective test set. We filtered out the images that were not classified correctly. We call the correctly classified images from the test set \textit{candidate} images. The number of candidates for each network are shown in \tabref{result_small_crown} and \tabref{result_big}.

For each candidate image $I_0$, we define the $L_\infty(I_0,\epsilon)$ based adversarial region by selecting challenging values of $\epsilon$ that are commonly used for testing the precision and scalability of verifiers in the literature. The $\epsilon$ values used for defining $L_\infty(I_0,\epsilon)$ for each neural network are shown in \tabref{result_small_crown} and \ref{Ta:result_big}. We used larger values of $\epsilon$ for testing DiffAI and \crown trained networks since these networks are more robust than the PGD and normally trained networks. However, DiffAI and \crown trained networks suffer from a substantial drop in test accuracy (see \#Candidates in \tabref{result_small_crown} and \tabref{result_big}). Furthermore, these networks are also easier to verify and thus even imprecise verifiers like \crown verify large number of properties on these networks while \tool takes advantage of the ease of verification by terminating the backsubstitution early, as explained in \secref{early_termination}. This leads to the runtimes of \tool being orders of magnitude smaller on DiffAI and \crown trained networks compared to normally trained or PGD trained networks. We also note that the $\epsilon$ values for MNIST based networks are larger than for CIFAR10 based networks for the same reason.

\subsection{Results on medium networks}
\textbf{Comparison with \crown.}
Table \ref{Ta:result_small_crown} compares the precision and the median runtime of \crown and \tool on the medium fully-connected and convolutional networks for 10,000 images.
We use the implementation of \crown publicly available from \cite{zhang2020towards}.
%
On normally trained networks, \crown does not prove any properties, while \tool proves $20,695$ overall. On DiffAI and \crown trained networks \tool proves an additional $8,863$ properties overall compared to \crown.
\crown is more precise on these networks than on normally-trained networks because inexact verifiers only sacrifice precision for scalability on neurons that are input to a ReLU and can take both positive and negative values during analysis. The number of such neurons for networks trained to be provable robust is relatively low. As can be seen, \tool improves upon the state-of-the-art results.
\crown is up to 45x faster than \tool on provably robust networks, and over 2,000x faster on normally trained networks. The speed of \crown comes at the cost of imprecision and the lack of floating point soundness guarantees. 

\textbf{Distribution of runtimes.}
The runtimes of \tool for normally and PGD trained networks are roughly normally distributed. On the other hand, the cumulative distribution function (CFD) of runtimes for DiffAI and \crown trained networks has a big tail of values which are orders of magnitudes larger than the median. This is because the early termination succeeds in the majority of cases for robustly trained networks yielding very small runtimes. In the small number of instances when it fails, the runtime is quite high. The CFD plots for all networks are in the appendix \ref{Se:cdf_appendix}.

\textbf{Comparison with \deeppoly.}
Table \ref{Ta:result_small_deeppoly} compares the precision and runtime of \deeppoly and \tool on six of our medium networks for the adversarial regions created on the first 500 test images on three MNIST and three CIFAR10 networks. The $\epsilon$ values are the same as in \tabref{result_small_crown}. While both have the same precision, \tool is up to 250x faster than \deeppoly on normally trained networks and up to 68,000x faster than \deeppoly on DiffAI trained networks.

\subsection{Results on big residual networks}
In \tabref{result_big} we compare the precision and runtime of \tool and \crown on our big residual networks, the largest being a \texttt{ResNet34} with almost 1M neurons. Since \crown does not support residual networks, we used our own implementation of \crown, which does not employ many of \crown's optimizations, such as batching, making it slower than the original, but equally precise.
%
\tool proves 1,299 samples for PGD trained networks overall while \crown cannot prove any. Furthermore \tool proves 420 additional properties compared to \crown on DiffAI trained networks. \tool only takes 34.5ms to verify our largest \texttt{ResNet34}.

\section{Conclusion} \label{Se:conclusion}

We presented a scalable neural network verifier, called \tool, for verifying the robustness of various types of deep neural networks on GPUs. \tool leverages GPU parallelization, sparsity in convolutional and residual networks, and an early termination mechanism. Our work advances the state-of-the-art by precisely verifying significantly larger CIFAR10 networks, with up to $1$M neurons, than possible with prior work. Based on our results, we believe that our work is a step in the direction towards scaling precise polyhedral analysis to even larger models.

\section{Acknowledgements}

We would like to thank Simon Schirm, Cheng Lai Low, and Marco Foco for their advice which helped shape the initial CUDA implementation of \tool.

\newpage

\nocite{langley00}

\bibliography{references}

\begin{thebibliography}{47}
\providecommand{\natexlab}[1]{#1}
\providecommand{\url}[1]{\texttt{#1}}
\expandafter\ifx\csname urlstyle\endcsname\relax
  \providecommand{\doi}[1]{doi: #1}\else
  \providecommand{\doi}{doi: \begingroup \urlstyle{rm}\Url}\fi

\bibitem[Abadi et~al.(2015)Abadi, Agarwal, Barham, Brevdo, Chen, Citro,
  Corrado, Davis, Dean, Devin, Ghemawat, Goodfellow, Harp, Irving, Isard, Jia,
  Jozefowicz, Kaiser, Kudlur, Levenberg, Man\'{e}, Monga, Moore, Murray, Olah,
  Schuster, Shlens, Steiner, Sutskever, Talwar, Tucker, Vanhoucke, Vasudevan,
  Vi\'{e}gas, Vinyals, Warden, Wattenberg, Wicke, Yu, and Zheng]{tensorflow:15}
Abadi, M., Agarwal, A., Barham, P., Brevdo, E., Chen, Z., Citro, C., Corrado,
  G.~S., Davis, A., Dean, J., Devin, M., Ghemawat, S., Goodfellow, I., Harp,
  A., Irving, G., Isard, M., Jia, Y., Jozefowicz, R., Kaiser, L., Kudlur, M.,
  Levenberg, J., Man\'{e}, D., Monga, R., Moore, S., Murray, D., Olah, C.,
  Schuster, M., Shlens, J., Steiner, B., Sutskever, I., Talwar, K., Tucker, P.,
  Vanhoucke, V., Vasudevan, V., Vi\'{e}gas, F., Vinyals, O., Warden, P.,
  Wattenberg, M., Wicke, M., Yu, Y., and Zheng, X.
\newblock {TensorFlow}: Large-scale machine learning on heterogeneous systems,
  2015.
\newblock URL \url{http://tensorflow.org/}.

\bibitem[Balunovic \& Vechev(2020)Balunovic and
  Vechev]{Balunovic2020Adversarial}
Balunovic, M. and Vechev, M.
\newblock Adversarial training and provable defenses: Bridging the gap.
\newblock In \emph{International Conference on Learning Representations
  (ICLR)}, 2020.

\bibitem[Balunovic et~al.(2019)Balunovic, Baader, Singh, Gehr, and
  Vechev]{BalunovicBSGV:19}
Balunovic, M., Baader, M., Singh, G., Gehr, T., and Vechev, M.~T.
\newblock Certifying geometric robustness of neural networks.
\newblock In \emph{Proc. Neural Information Processing Systems (NeurIPS)}, pp.\
   15287--15297, 2019.

\bibitem[Boopathy et~al.(2019)Boopathy, Weng, Chen, Liu, and
  Daniel]{cnncert:19}
Boopathy, A., Weng, T.-W., Chen, P.-Y., Liu, S., and Daniel, L.
\newblock Cnn-cert: An efficient framework for certifying robustness of
  convolutional neural networks.
\newblock In \emph{Proc. AAAI Conference on Artificial Intelligence (AAAI)},
  Jan 2019.

\bibitem[Bunel et~al.(2018)Bunel, Turkaslan, Torr, Kohli, and Kumar]{bunel:18}
Bunel, R., Turkaslan, I., Torr, P.~H., Kohli, P., and Kumar, M.~P.
\newblock A unified view of piecewise linear neural network verification.
\newblock In \emph{Proc. Advances in Neural Information Processing Systems
  (NeurIPS)}, pp.\  4795--4804, 2018.

\bibitem[Carlini \& Wagner(2017)Carlini and Wagner]{Carlini:17}
Carlini, N. and Wagner, D.~A.
\newblock Towards evaluating the robustness of neural networks.
\newblock In \emph{Proc. {IEEE} Symposium on Security and Privacy ({SP})}, pp.\
   39--57, 2017.

\bibitem[Dathathri et~al.(2020)Dathathri, Dvijotham, Kurakin, Raghunathan,
  Uesato, Bunel, Shankar, Steinhardt, Goodfellow, Liang, and
  Kohli]{DathathriDKRUBS:20}
Dathathri, S., Dvijotham, K., Kurakin, A., Raghunathan, A., Uesato, J., Bunel,
  R., Shankar, S., Steinhardt, J., Goodfellow, I.~J., Liang, P., and Kohli, P.
\newblock Enabling certification of verification-agnostic networks via
  memory-efficient semidefinite programming.
\newblock In \emph{Proc. Neural Information Processing Systems (NeurIPS)},
  2020.

\bibitem[Dong et~al.(2018)Dong, Liao, Pang, Su, Zhu, Hu, and Li]{dong:18}
Dong, Y., Liao, F., Pang, T., Su, H., Zhu, J., Hu, X., and Li, J.
\newblock Boosting adversarial attacks with momentum.
\newblock In \emph{Proc. Computer Vision and Pattern Recognition (CVPR)}, 2018.

\bibitem[Dvijotham et~al.(2018)Dvijotham, Stanforth, Gowal, Mann, and
  Kohli]{Dvijotham:18}
Dvijotham, K., Stanforth, R., Gowal, S., Mann, T., and Kohli, P.
\newblock A dual approach to scalable verification of deep networks.
\newblock In \emph{Proc. Uncertainty in Artificial Intelligence (UAI)}, pp.\
  162--171, 2018.

\bibitem[Ehlers(2017)]{Ehlers:17}
Ehlers, R.
\newblock Formal verification of piece-wise linear feed-forward neural
  networks.
\newblock In \emph{Automated Technology for Verification and Analysis (ATVA)},
  2017.

\bibitem[Gehr et~al.(2018)Gehr, Mirman, Drachsler-Cohen, Tsankov, Chaudhuri,
  and Vechev]{Gehr:18}
Gehr, T., Mirman, M., Drachsler-Cohen, D., Tsankov, P., Chaudhuri, S., and
  Vechev, M.
\newblock {AI2}: Safety and robustness certification of neural networks with
  abstract interpretation.
\newblock In \emph{Proc. {IEEE} Symposium on Security and Privacy (SP)},
  volume~00, pp.\  948--963, 2018.

\bibitem[Gowal et~al.(2018)Gowal, Dvijotham, Stanforth, Bunel, Qin, Uesato,
  Arandjelovic, Mann, and Kohli]{IBP:18}
Gowal, S., Dvijotham, K., Stanforth, R., Bunel, R., Qin, C., Uesato, J.,
  Arandjelovic, R., Mann, T.~A., and Kohli, P.
\newblock On the effectiveness of interval bound propagation for training
  verifiably robust models.
\newblock \emph{CoRR}, abs/1810.12715, 2018.

\bibitem[Grund(1982)]{autodiff}
Grund, F.
\newblock Rall, louis b., automatic differentiation: Techniques and
  applications. lecture notes in computer science 120.
\newblock \emph{ZAMM - Journal of Applied Mathematics and Mechanics},
  62\penalty0 (7), 1982.

\bibitem[Jia \& Rinard(2020)Jia and Rinard]{jia2020exploiting}
Jia, K. and Rinard, M.
\newblock Exploiting verified neural networks via floating point numerical
  error, 2020.

\bibitem[Katz et~al.(2017)Katz, Barrett, Dill, Julian, and
  Kochenderfer]{katz:17}
Katz, G., Barrett, C.~W., Dill, D.~L., Julian, K., and Kochenderfer, M.~J.
\newblock Reluplex: An efficient {SMT} solver for verifying deep neural
  networks.
\newblock In \emph{Computer Aided Verification - 29th International Conference,
  {CAV} 2017, Heidelberg, Germany, July 24-28, 2017, Proceedings, Part {I}},
  2017.

\bibitem[Katz et~al.(2019)Katz, Huang, Ibeling, Julian, Lazarus, Lim, Shah,
  Thakoor, Wu, Zelji{\'{c}}, Dill, Kochenderfer, and Barrett]{Marabou:19}
Katz, G., Huang, D.~A., Ibeling, D., Julian, K., Lazarus, C., Lim, R., Shah,
  P., Thakoor, S., Wu, H., Zelji{\'{c}}, A., Dill, D.~L., Kochenderfer, M.~J.,
  and Barrett, C.
\newblock The marabou framework for verification and analysis of deep neural
  networks.
\newblock In \emph{Proc. Computer Aided Verification (CAV)}, pp.\  443--452,
  2019.

\bibitem[Krizhevsky(2009)]{Krizhevsky:09}
Krizhevsky, A.
\newblock Learning multiple layers of features from tiny images.
\newblock Technical report, 2009.

\bibitem[Lecun et~al.(1998)Lecun, Bottou, Bengio, and Haffner]{Lecun:98}
Lecun, Y., Bottou, L., Bengio, Y., and Haffner, P.
\newblock Gradient-based learning applied to document recognition.
\newblock In \emph{Proc. of the {IEEE}}, pp.\  2278--2324, 1998.

\bibitem[Madry et~al.(2018)Madry, Makelov, Schmidt, Tsipras, and
  Vladu]{madry:17}
Madry, A., Makelov, A., Schmidt, L., Tsipras, D., and Vladu, A.
\newblock Towards deep learning models resistant to adversarial attacks.
\newblock In \emph{Proc. International Conference on Learning Representations
  (ICLR)}, 2018.

\bibitem[Miné(2004)]{rounding}
Miné, A.
\newblock Relational abstract domains for the detection of floating-point
  run-time errors.
\newblock In \emph{Proc. EuropeanSymposium on Programming (ESOP)}, 2004.

\bibitem[Mirman et~al.(2018)Mirman, Gehr, and Vechev]{mirman:18}
Mirman, M., Gehr, T., and Vechev, M.
\newblock Differentiable abstract interpretation for provably robust neural
  networks.
\newblock In \emph{Proc. International Conference on Machine Learning
  ({ICML})}, pp.\  3575--3583, 2018.

\bibitem[Mirman et~al.(2019)Mirman, Singh, and Vechev]{mirman:19}
Mirman, M., Singh, G., and Vechev, M.~T.
\newblock A provable defense for deep residual networks.
\newblock \emph{CoRR}, abs/1903.12519, 2019.

\bibitem[Mirman et~al.(2020)Mirman, Gehr, and Vechev]{mirman:20}
Mirman, M., Gehr, T., and Vechev, M.
\newblock Robustness certification of generative models, 2020.

\bibitem[Paszke et~al.(2017)Paszke, Gross, Chintala, Chanan, Yang, DeVito, Lin,
  Desmaison, Antiga, and Lerer]{pytorch:17}
Paszke, A., Gross, S., Chintala, S., Chanan, G., Yang, E., DeVito, Z., Lin, Z.,
  Desmaison, A., Antiga, L., and Lerer, A.
\newblock Automatic differentiation in pytorch.
\newblock 2017.

\bibitem[Paterson et~al.(2021)Paterson, Wu, Grese, Calinescu, Pasareanu, and
  Barrett]{paterson:21}
Paterson, C., Wu, H., Grese, J., Calinescu, R., Pasareanu, C.~S., and Barrett,
  C.
\newblock Deepcert: Verification of contextually relevant robustness for neural
  network image classifiers, 2021.

\bibitem[Pei et~al.(2017)Pei, Cao, Yang, and Jana]{Pei:17}
Pei, K., Cao, Y., Yang, J., and Jana, S.
\newblock Deepxplore: Automated whitebox testing of deep learning systems.
\newblock In \emph{Proc. Symposium on Operating Systems Principles (SOSP)},
  pp.\  1--18, 2017.

\bibitem[Raghunathan et~al.(2018)Raghunathan, Steinhardt, and Liang]{aditi:18}
Raghunathan, A., Steinhardt, J., and Liang, P.~S.
\newblock Semidefinite relaxations for certifying robustness to adversarial
  examples.
\newblock In \emph{Advances in Neural Information Processing Systems
  (NeurIPS)}, pp.\  10877--10887. 2018.

\bibitem[Ruan et~al.(2018)Ruan, Huang, and Kwiatkowska]{ruanhk:18}
Ruan, W., Huang, X., and Kwiatkowska, M.
\newblock Reachability analysis of deep neural networks with provable
  guarantees.
\newblock In \emph{Proc. International Joint Conference on Artificial
  Intelligence, (IJCAI)}, 2018.

\bibitem[Ruoss et~al.(2020)Ruoss, Balunovic, Fischer, and Vechev]{RuossBFV:20}
Ruoss, A., Balunovic, M., Fischer, M., and Vechev, M.~T.
\newblock Learning certified individually fair representations.
\newblock In \emph{Proc. Neural Information Processing Systems (NeurIPS)},
  2020.

\bibitem[Ruoss et~al.(2021)Ruoss, Baader, Balunovic, and Vechev]{Spatial:21}
Ruoss, A., Baader, M., Balunovic, M., and Vechev, M.~T.
\newblock Efficient certification of spatial robustness.
\newblock 2021.

\bibitem[Salman et~al.(2019)Salman, Yang, Zhang, Hsieh, and Zhang]{salman:19}
Salman, H., Yang, G., Zhang, H., Hsieh, C., and Zhang, P.
\newblock A convex relaxation barrier to tight robustness verification of
  neural networks.
\newblock \emph{CoRR}, abs/1902.08722, 2019.

\bibitem[Singh et~al.(2018)Singh, Gehr, Mirman, P\"{u}schel, and
  Vechev]{Singh:18}
Singh, G., Gehr, T., Mirman, M., P\"{u}schel, M., and Vechev, M.
\newblock Fast and effective robustness certification.
\newblock In \emph{Proc. Advances in Neural Information Processing Systems
  (NeurIPS)}, pp.\  10825--10836. 2018.

\bibitem[Singh et~al.(2019{\natexlab{a}})Singh, Ganvir, Püschel, and
  Vechev]{singh2019krelu}
Singh, G., Ganvir, R., Püschel, M., and Vechev, M.
\newblock Beyond the single neuron convex barrier for neural network
  certification.
\newblock In \emph{Advances in Neural Information Processing Systems
  (NeurIPS)}. 2019{\natexlab{a}}.

\bibitem[Singh et~al.(2019{\natexlab{b}})Singh, Gehr, P\"{u}schel, and
  Vechev]{Singh:19}
Singh, G., Gehr, T., P\"{u}schel, M., and Vechev, M.
\newblock An abstract domain for certifying neural networks.
\newblock \emph{Proc. ACM Program. Lang.}, 3\penalty0 (POPL):\penalty0
  41:1--41:30, 2019{\natexlab{b}}.
\newblock ISSN 2475-1421.

\bibitem[Singh et~al.(2019{\natexlab{c}})Singh, Gehr, Püschel, and
  Vechev]{singhrobustness:19}
Singh, G., Gehr, T., Püschel, M., and Vechev, M.
\newblock Boosting robustness certification of neural networks.
\newblock In \emph{International Conference on Learning Representations
  (ICLR)}, 2019{\natexlab{c}}.

\bibitem[Szegedy et~al.(2013)Szegedy, Zaremba, Sutskever, Bruna, Erhan,
  Goodfellow, and Fergus]{szegedy:13}
Szegedy, C., Zaremba, W., Sutskever, I., Bruna, J., Erhan, D., Goodfellow, I.,
  and Fergus, R.
\newblock Intriguing properties of neural networks.
\newblock \emph{arXiv preprint arXiv:1312.6199}, 2013.

\bibitem[Tjandraatmadja et~al.(2020)Tjandraatmadja, Anderson, Huchette, Ma,
  Patel, and Vielma]{Tjandraatmadja:20}
Tjandraatmadja, C., Anderson, R., Huchette, J., Ma, W., Patel, K., and Vielma,
  J.~P.
\newblock The convex relaxation barrier, revisited: Tightened single-neuron
  relaxations for neural network verification.
\newblock In \emph{Proc. Neural Information Processing Systems (NeurIPS)},
  2020.

\bibitem[Tjeng et~al.(2019)Tjeng, Xiao, and Tedrake]{tjeng:18}
Tjeng, V., Xiao, K.~Y., and Tedrake, R.
\newblock Evaluating robustness of neural networks with mixed integer
  programming.
\newblock In \emph{International Conference on Learning Representations,
  (ICLR)}, 2019.

\bibitem[Tran et~al.(2020)Tran, Bak, Xiang, and Johnson]{TranBXJ:20}
Tran, H., Bak, S., Xiang, W., and Johnson, T.~T.
\newblock Verification of deep convolutional neural networks using imagestars.
\newblock In Lahiri, S.~K. and Wang, C. (eds.), \emph{Proc. Computer Aided
  Verification (CAV)}, volume 12224 of \emph{Lecture Notes in Computer
  Science}, pp.\  18--42, 2020.

\bibitem[Wang et~al.()Wang, Pei, Whitehouse, Yang, and Jana]{Reluval:2018}
Wang, S., Pei, K., Whitehouse, J., Yang, J., and Jana, S.
\newblock Formal security analysis of neural networks using symbolic intervals.
\newblock In \emph{Proc. USENIX Conference on Security Symposium}, SEC'18, pp.\
   1599--1614.

\bibitem[Wang et~al.(2018)Wang, Pei, Whitehouse, Yang, and Jana]{shiqi:18}
Wang, S., Pei, K., Whitehouse, J., Yang, J., and Jana, S.
\newblock Efficient formal safety analysis of neural networks.
\newblock In \emph{Proc. Advances in Neural Information Processing Systems
  (NeurIPS)}, pp.\  6369--6379. 2018.

\bibitem[Weng et~al.(2018)Weng, Zhang, Chen, Song, Hsieh, Daniel, Boning, and
  Dhillon]{Weng:18}
Weng, L., Zhang, H., Chen, H., Song, Z., Hsieh, C.-J., Daniel, L., Boning, D.,
  and Dhillon, I.
\newblock Towards fast computation of certified robustness for {R}e{LU}
  networks.
\newblock In \emph{Proc. International Conference on Machine Learning (ICML)},
  pp.\  5276--5285, 2018.

\bibitem[Wong \& Kolter(2018)Wong and Kolter]{wong18a}
Wong, E. and Kolter, Z.
\newblock Provable defenses against adversarial examples via the convex outer
  adversarial polytope.
\newblock In \emph{Proc. International Conference on Machine Learning (ICML)},
  pp.\  5286--5295, 2018.

\bibitem[Xu et~al.(2020)Xu, Shi, Zhang, Wang, Chang, Huang, Kailkhura, Lin, and
  Hsieh]{Lirpa:20}
Xu, K., Shi, Z., Zhang, H., Wang, Y., Chang, K., Huang, M., Kailkhura, B., Lin,
  X., and Hsieh, C.
\newblock Automatic perturbation analysis for scalable certified robustness and
  beyond.
\newblock 2020.

\bibitem[Zhang et~al.(2018)Zhang, Weng, Chen, Hsieh, and Daniel]{zhang:18}
Zhang, H., Weng, T.-W., Chen, P.-Y., Hsieh, C.-J., and Daniel, L.
\newblock Efficient neural network robustness certification with general
  activation functions.
\newblock In \emph{Proc. Advances in Neural Information Processing Systems
  (NeurIPS)}. 2018.

\bibitem[Zhang et~al.(2020)Zhang, Chen, Xiao, Gowal, Stanforth, Li, Boning, and
  Hsieh]{zhang2020towards}
Zhang, H., Chen, H., Xiao, C., Gowal, S., Stanforth, R., Li, B., Boning, D.,
  and Hsieh, C.-J.
\newblock Towards stable and efficient training of verifiably robust neural
  networks.
\newblock In \emph{International Conference on Learning Representations
  (ICLR)}, 2020.

\bibitem[Zombori et~al.(2021)Zombori, B{\'a}nhelyi, Csendes, Megyeri, and
  Jelasity]{zombori:21}
Zombori, D., B{\'a}nhelyi, B., Csendes, T., Megyeri, I., and Jelasity, M.
\newblock Fooling a complete neural network verifier.
\newblock In \emph{Proc. International Conference on Learning Representations
  (ICLR)}, 2021.

\end{thebibliography}
\bibliographystyle{mlsys2020}


\appendix
\section{Cumulative distribution functions of runtimes} \label{Se:cdf_appendix}
\begin{figure*}\label{fig:cdf}
\subfigure[ConvBig\_cifar10]{\includegraphics[width=0.5\columnwidth]{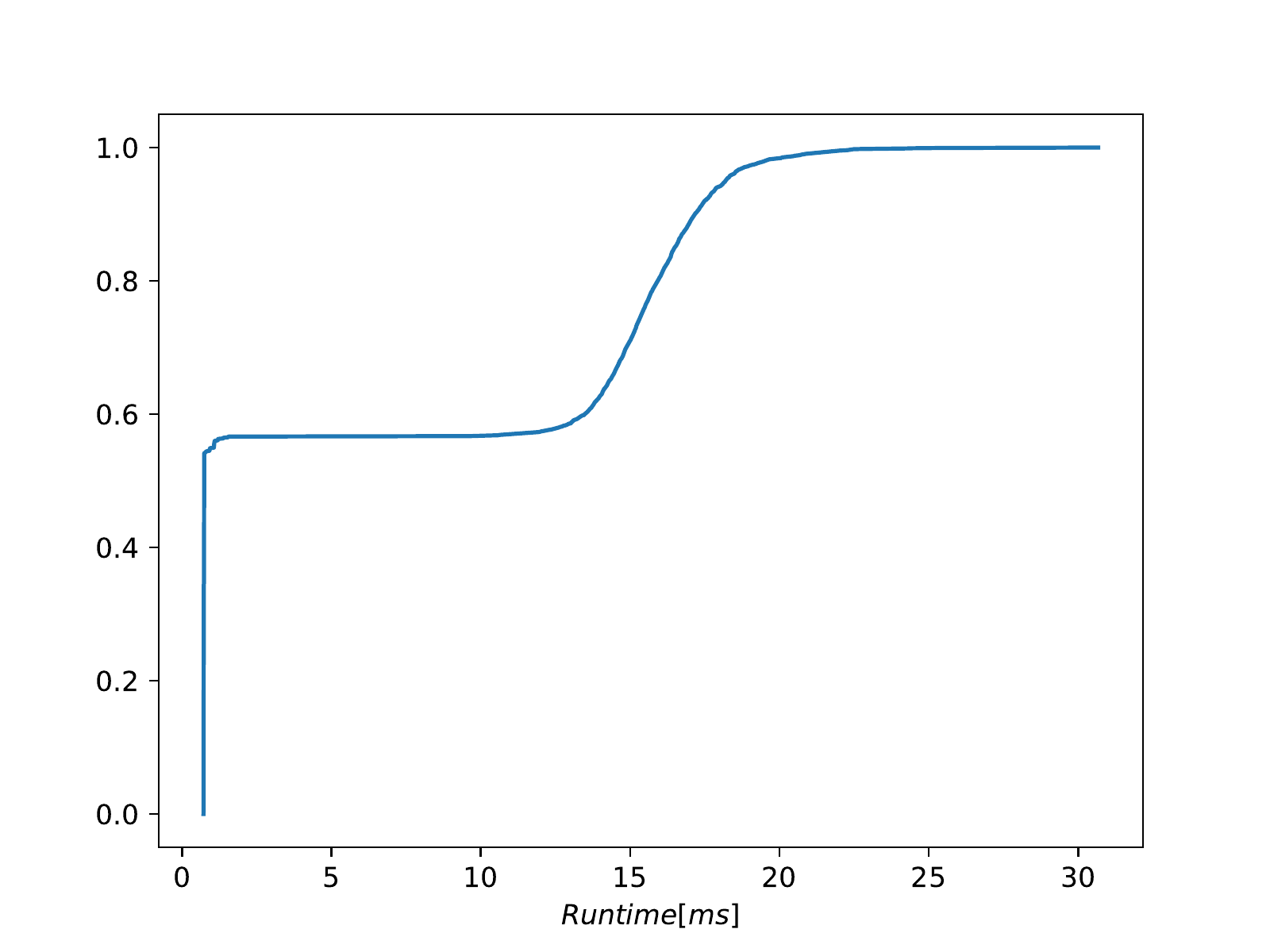}}
\subfigure[ConvBig\_mnist]{\includegraphics[width=0.5\columnwidth]{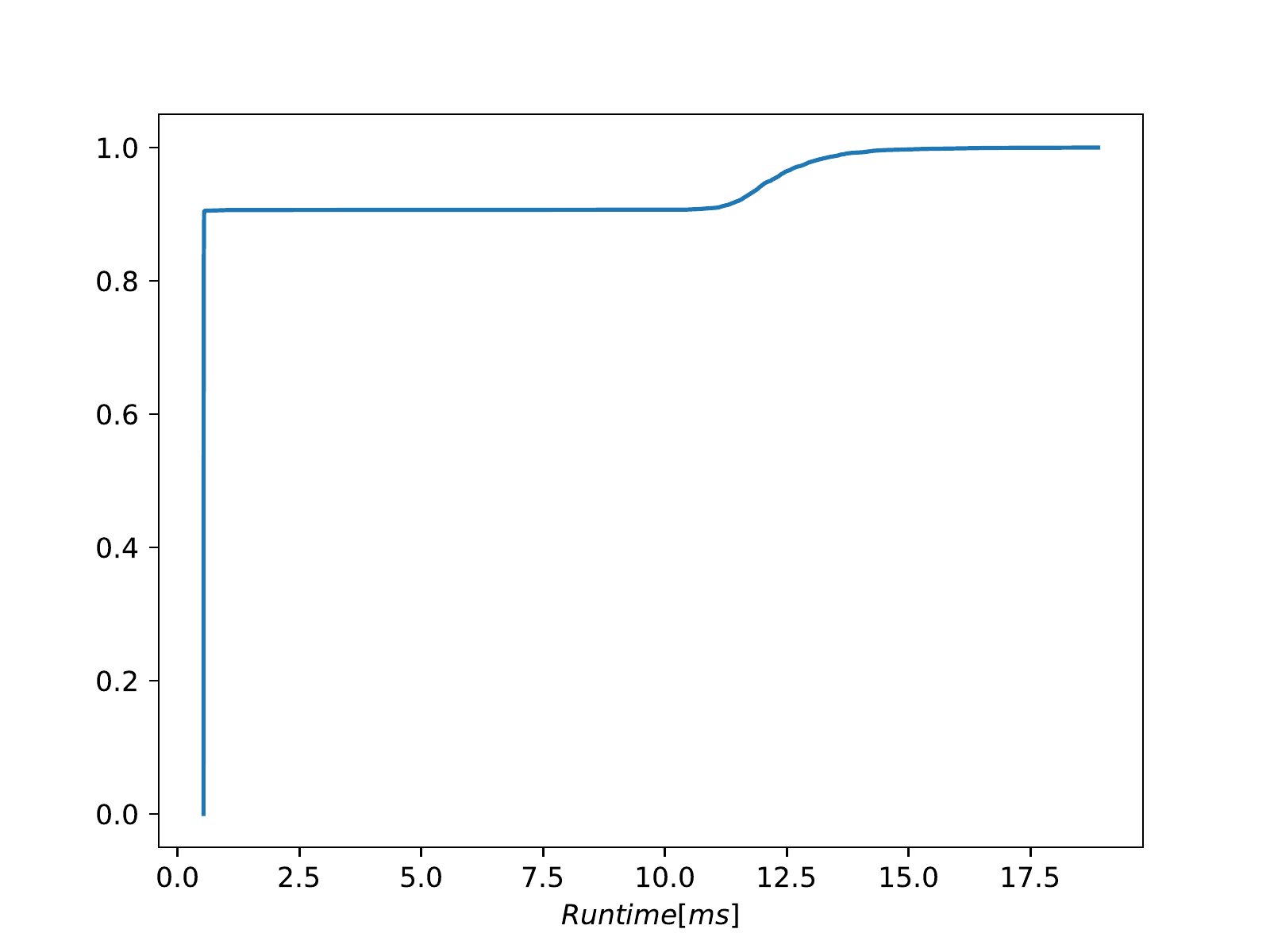}}
\subfigure[ConvLargeIBP]{\includegraphics[width=0.5\columnwidth]{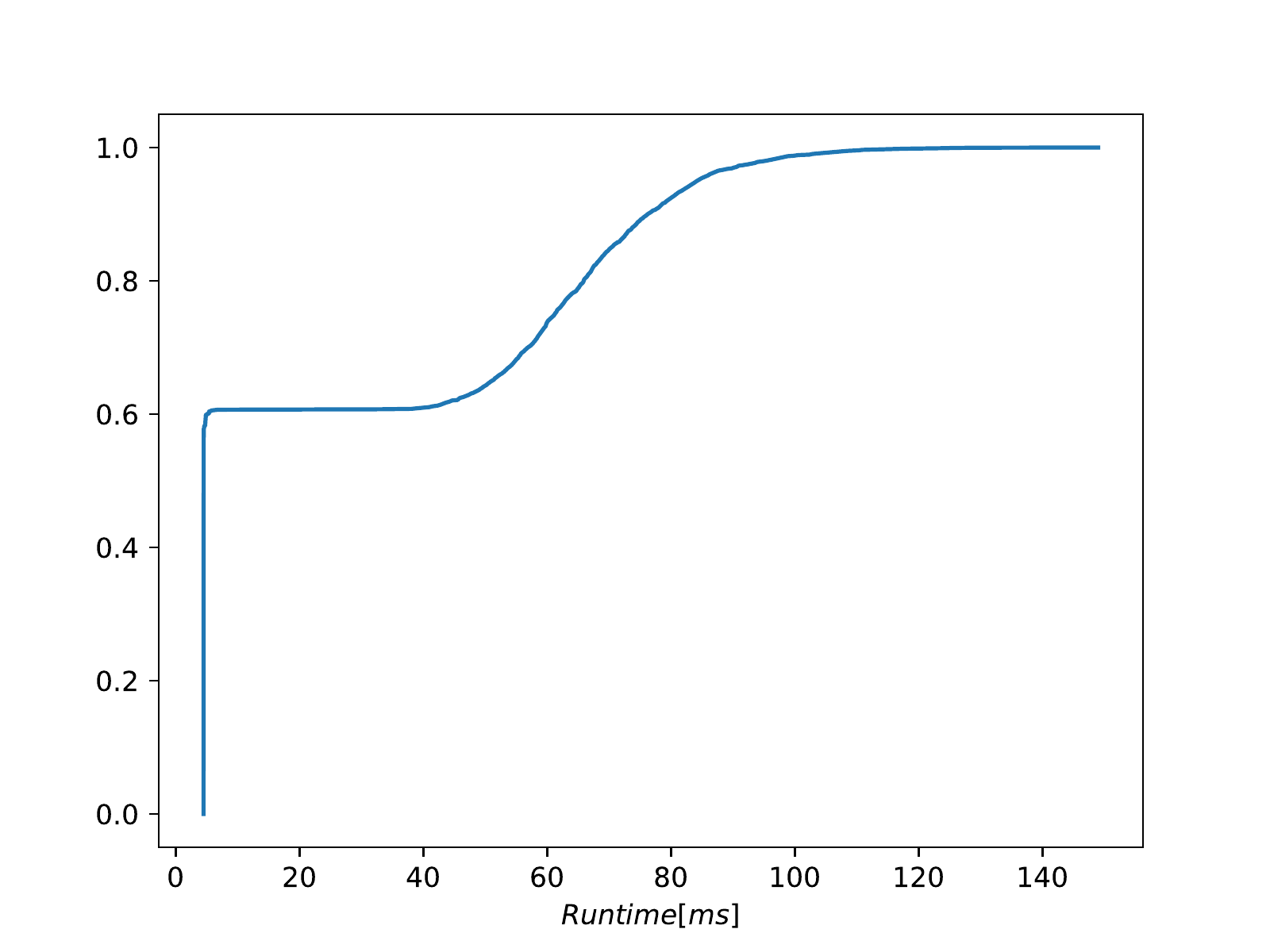}}
\subfigure[FFNN\_cifar10]{\includegraphics[width=0.5\columnwidth]{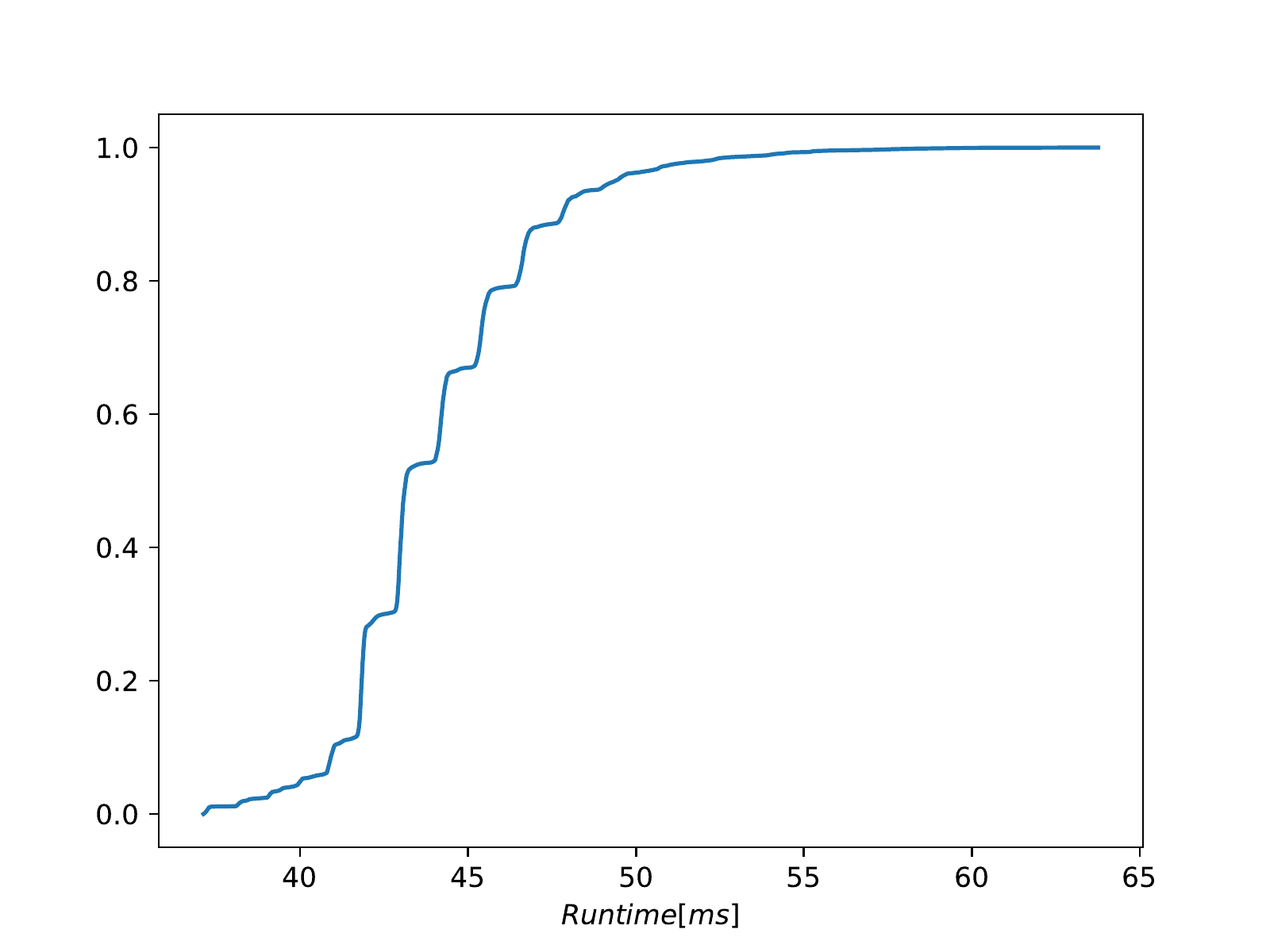}}\\
\subfigure[FFNN\_mnist]{\includegraphics[width=0.5\columnwidth]{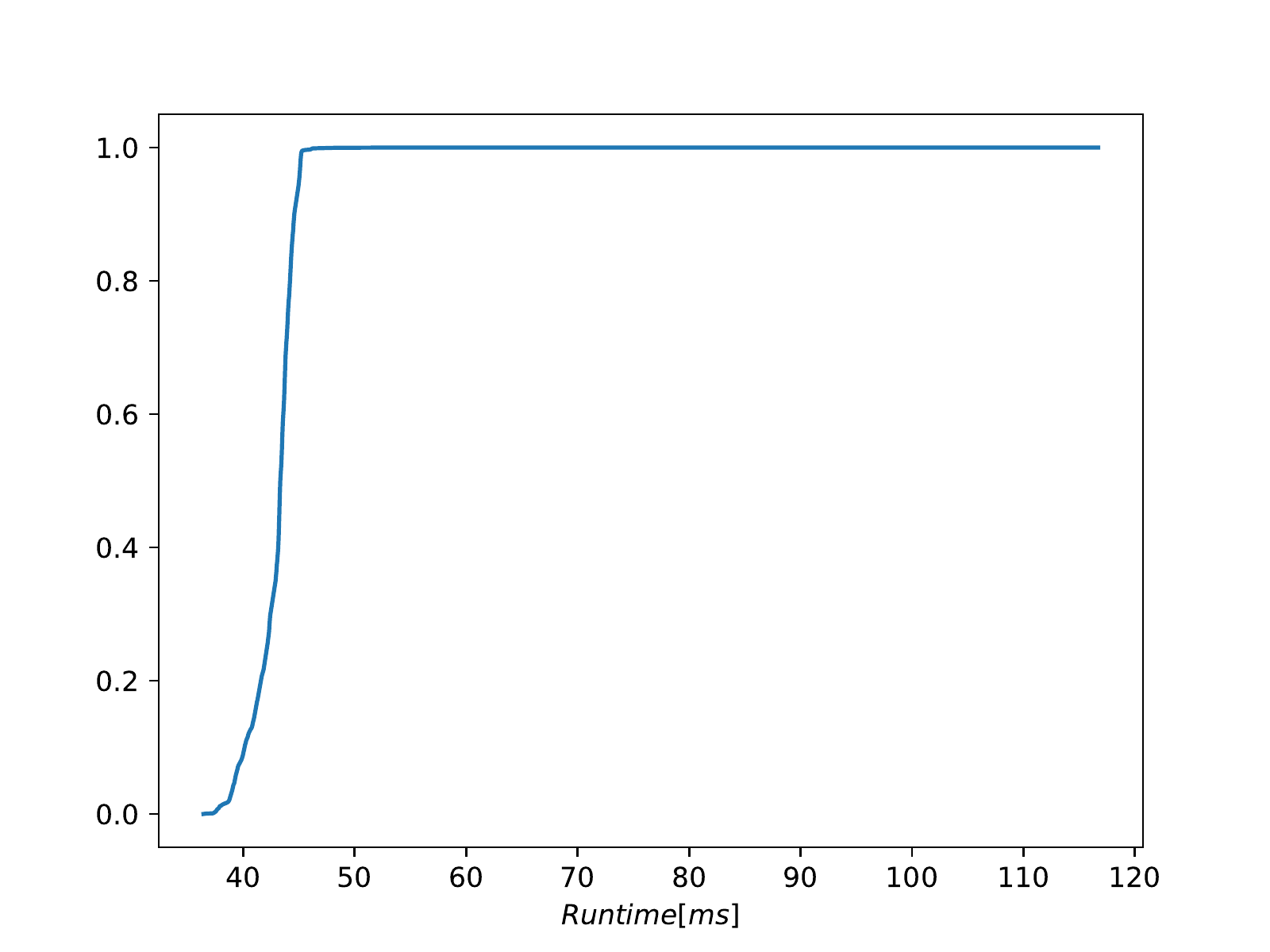}}
\subfigure[ResNet18\_DiffAI]{\includegraphics[width=0.5\columnwidth]{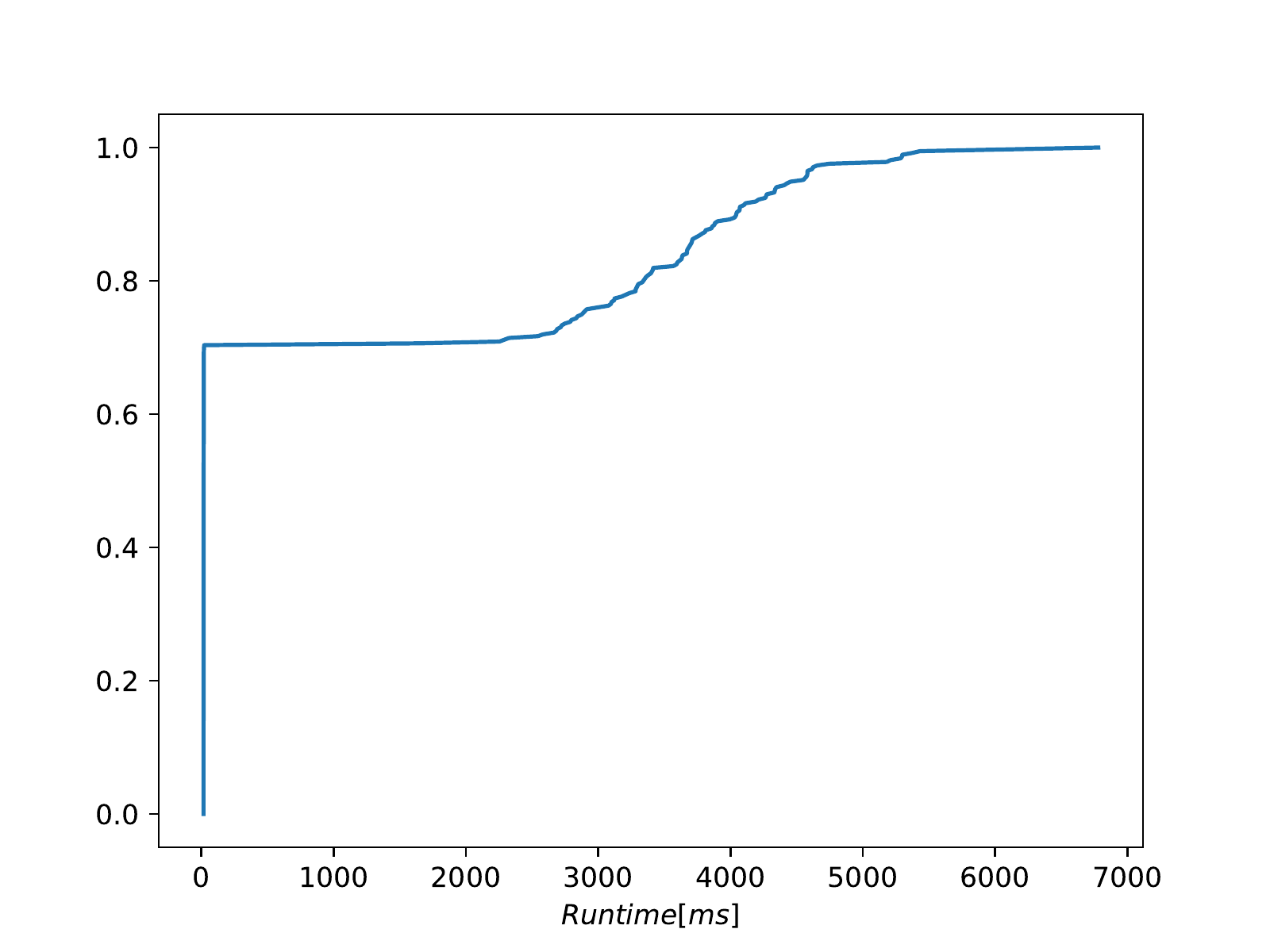}}
\subfigure[ResNet18\_PGD]{\includegraphics[width=0.5\columnwidth]{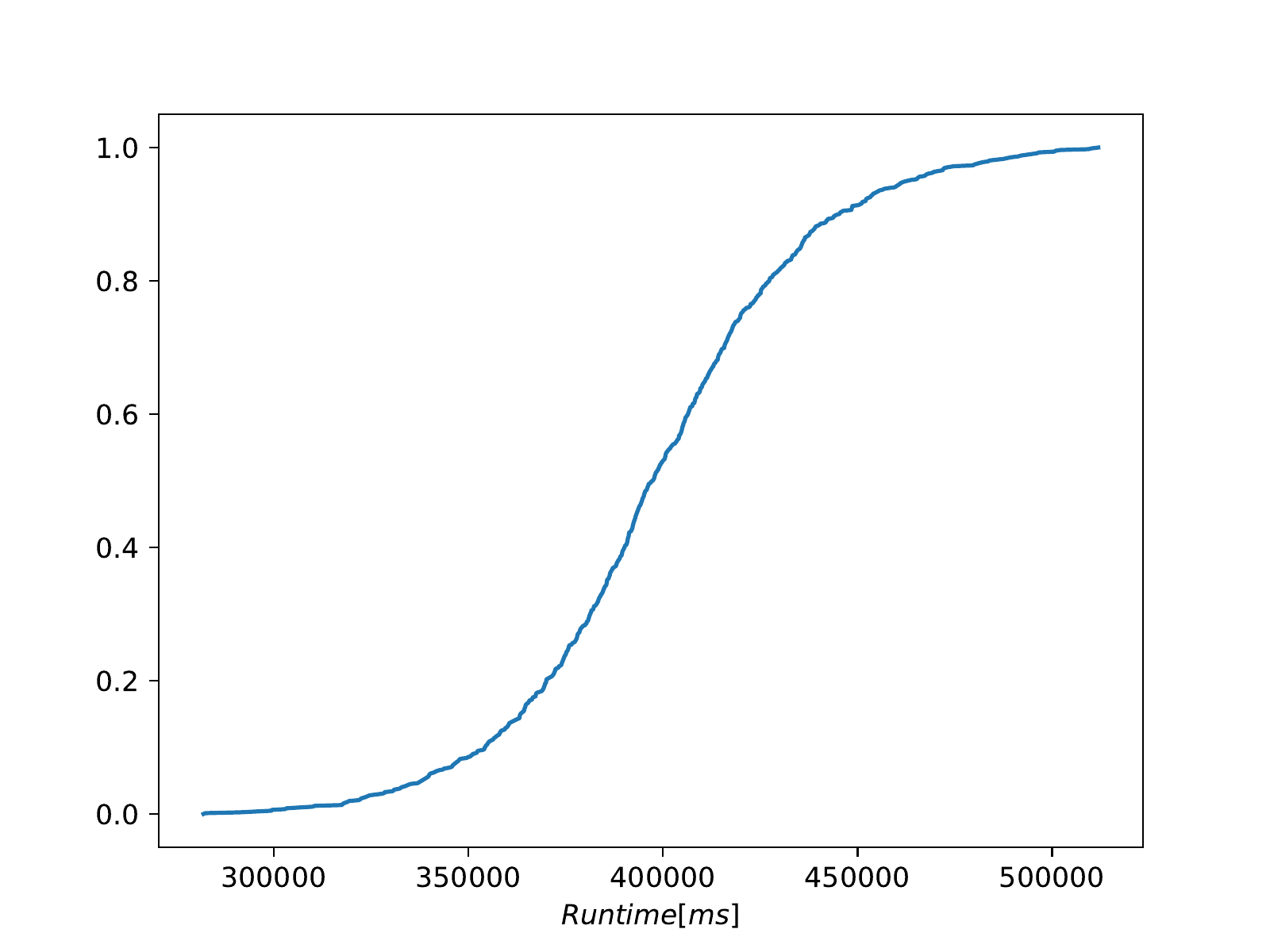}}
\subfigure[ResNet34\_DiffAI]{\includegraphics[width=0.5\columnwidth]{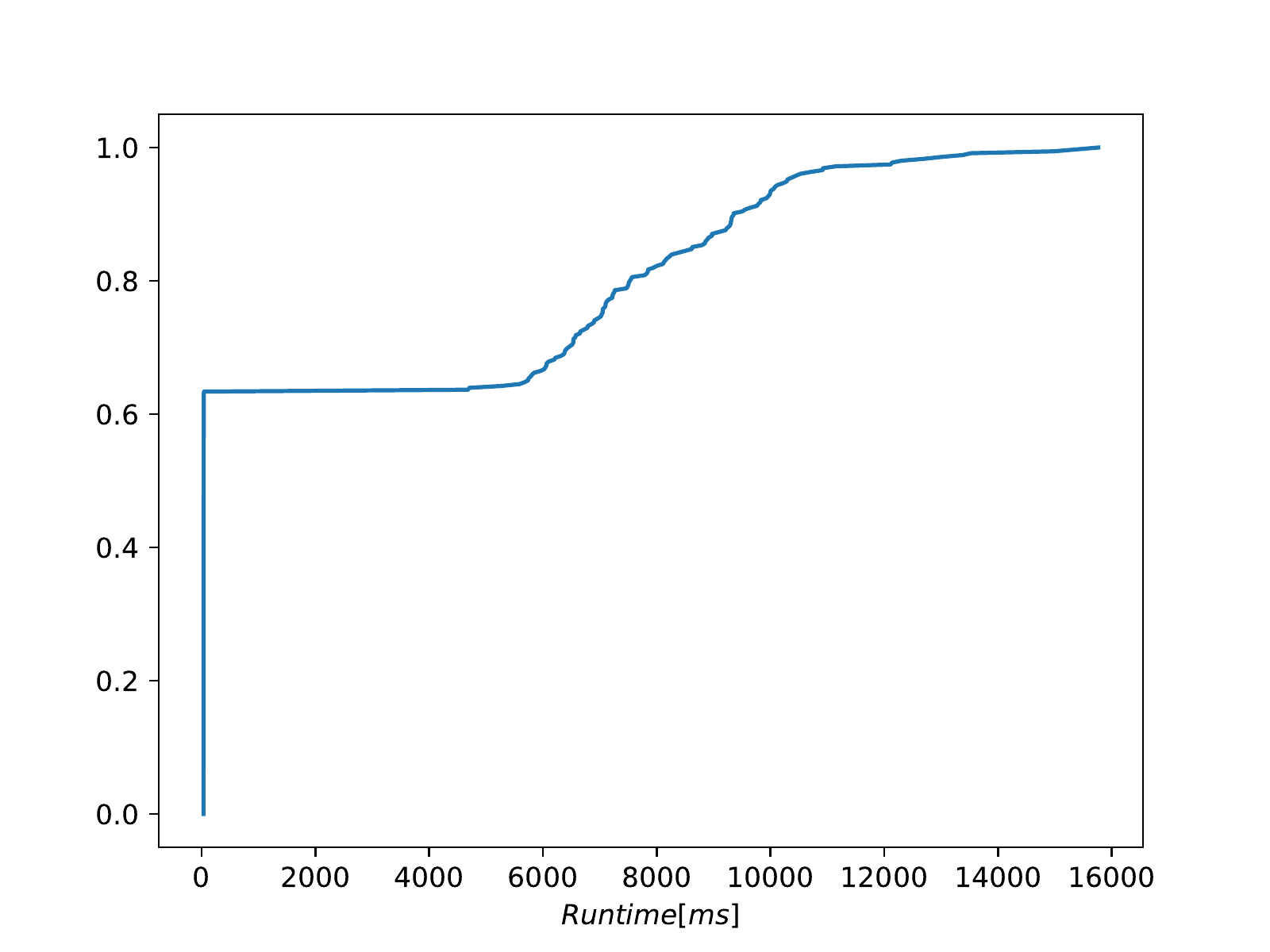}}\\
\subfigure[ResNetTiny\_DiffAI]{\includegraphics[width=0.5\columnwidth]{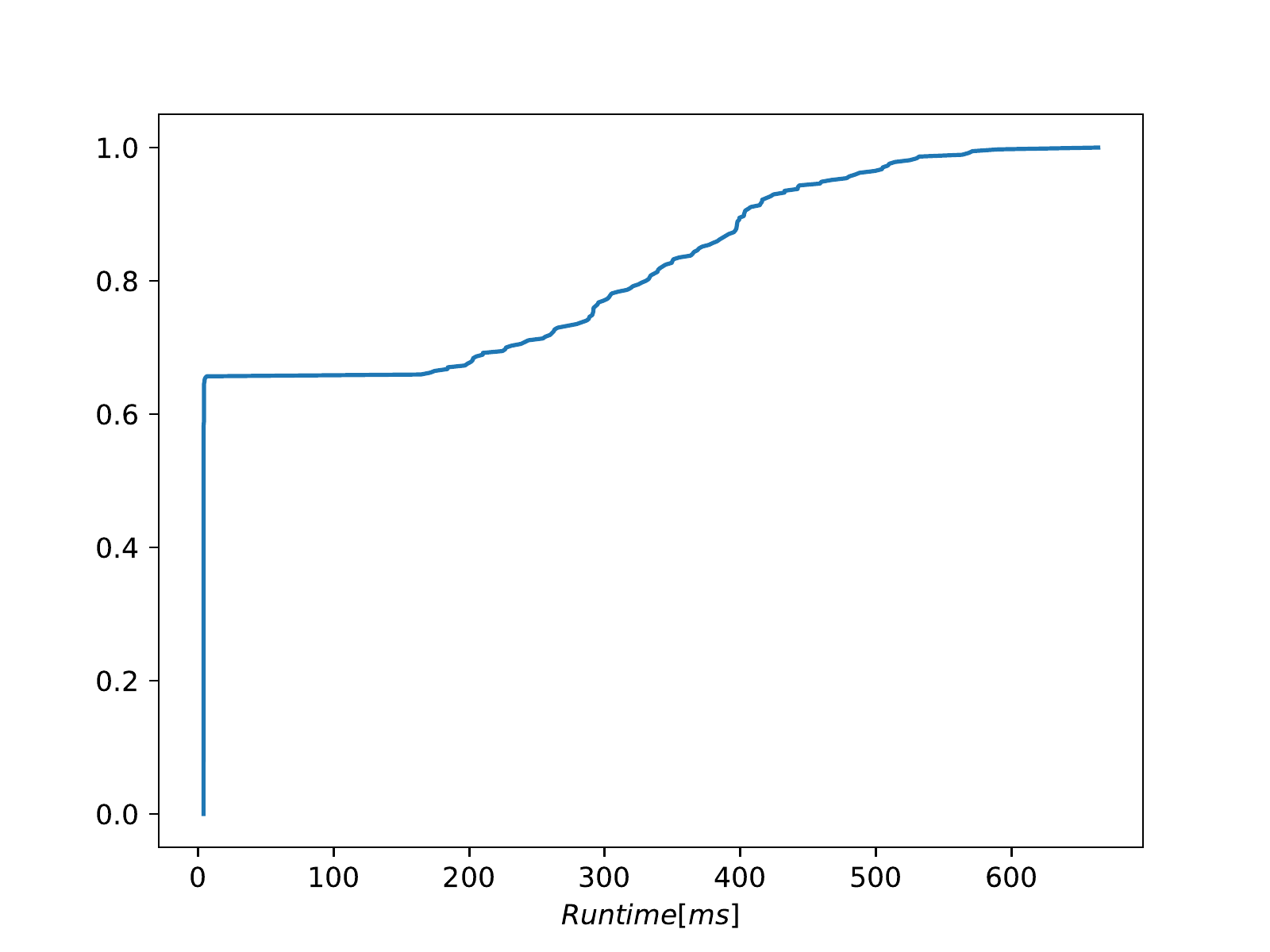}}
\subfigure[ResNetTiny\_PGD]{\includegraphics[width=0.5\columnwidth]{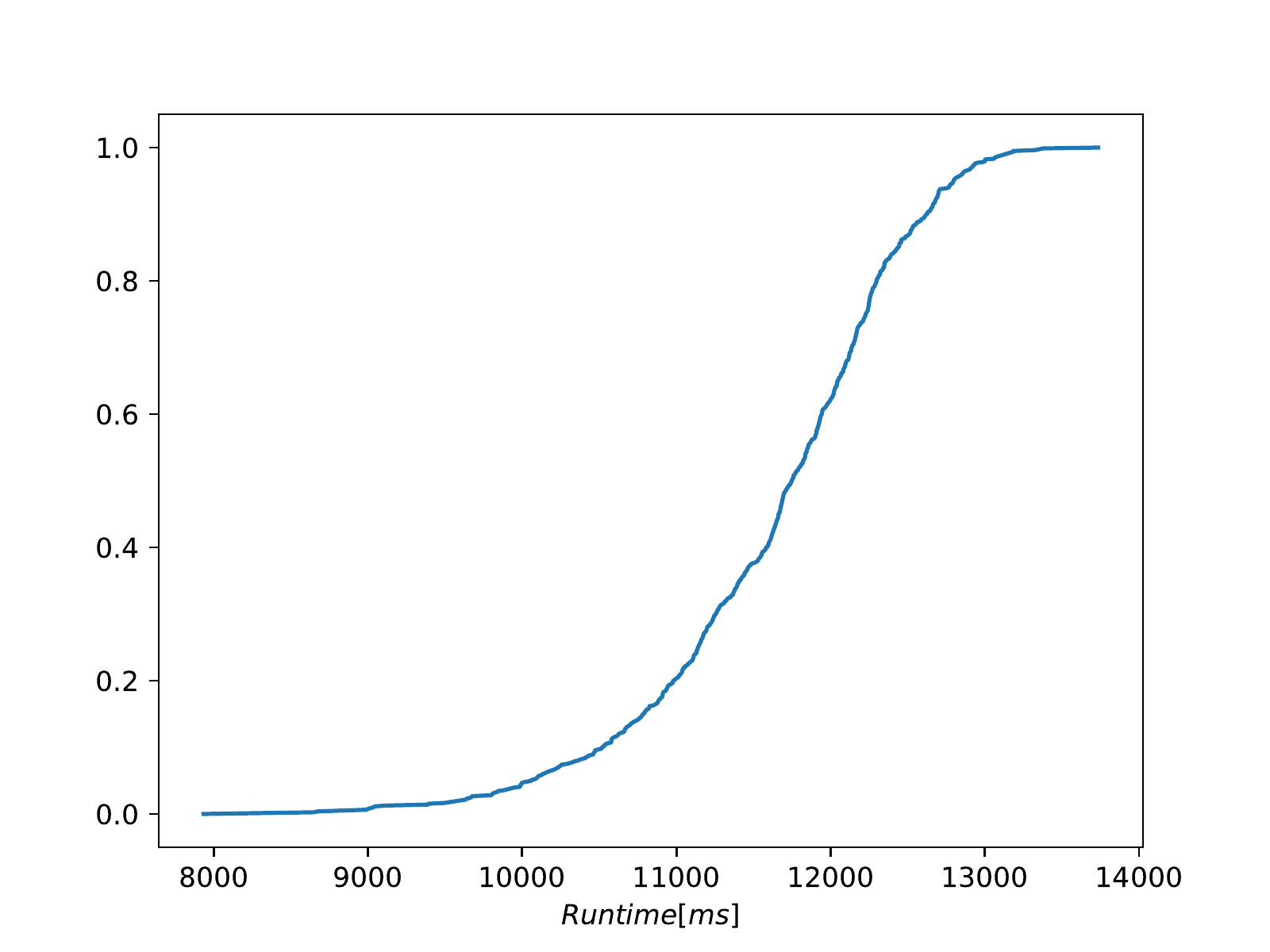}}
\subfigure[SkipNet18\_DiffAI]{\includegraphics[width=0.5\columnwidth]{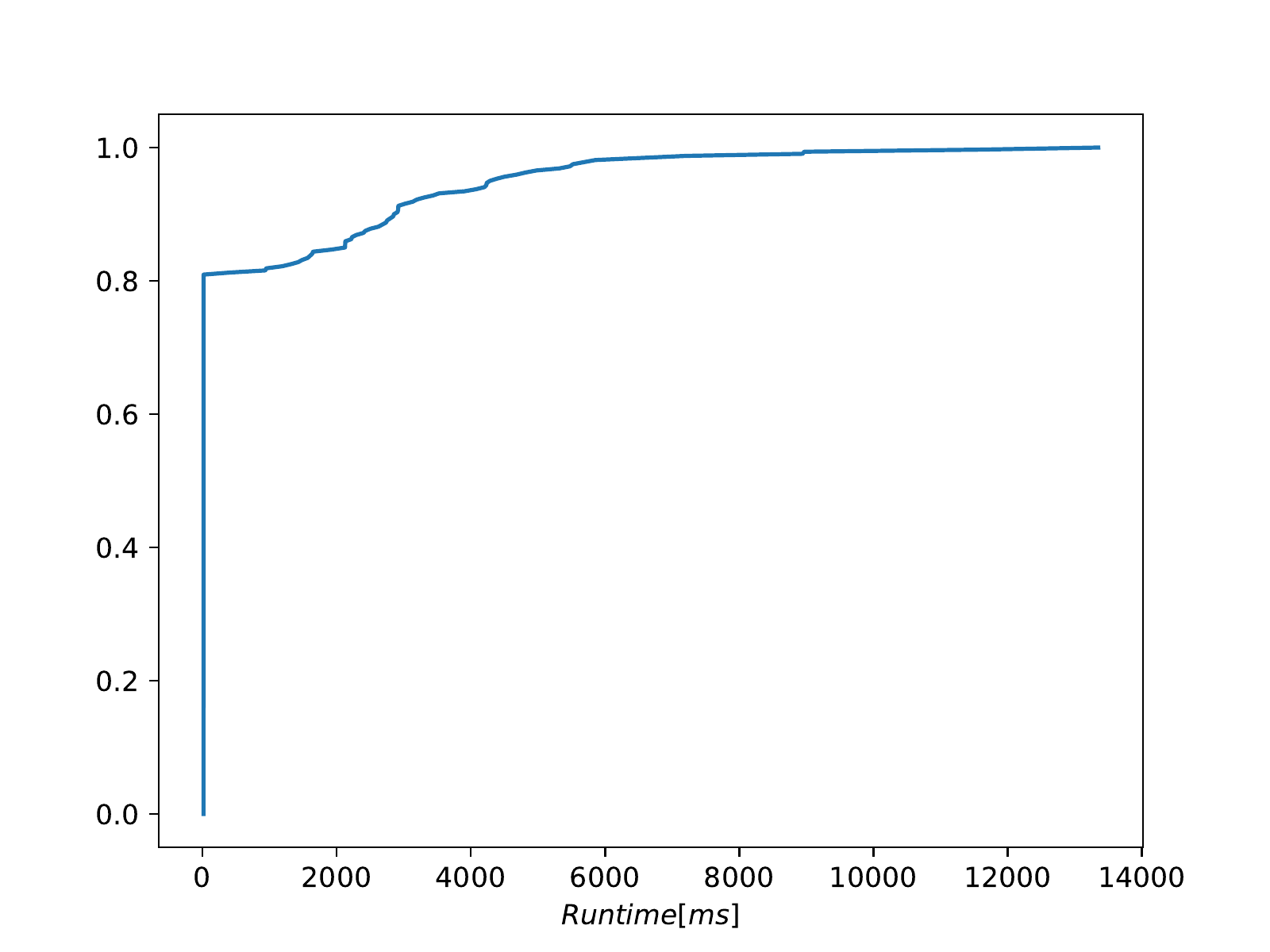}}
\subfigure[cifar\_crown\_large\_2\_255]{\includegraphics[width=0.5\columnwidth]{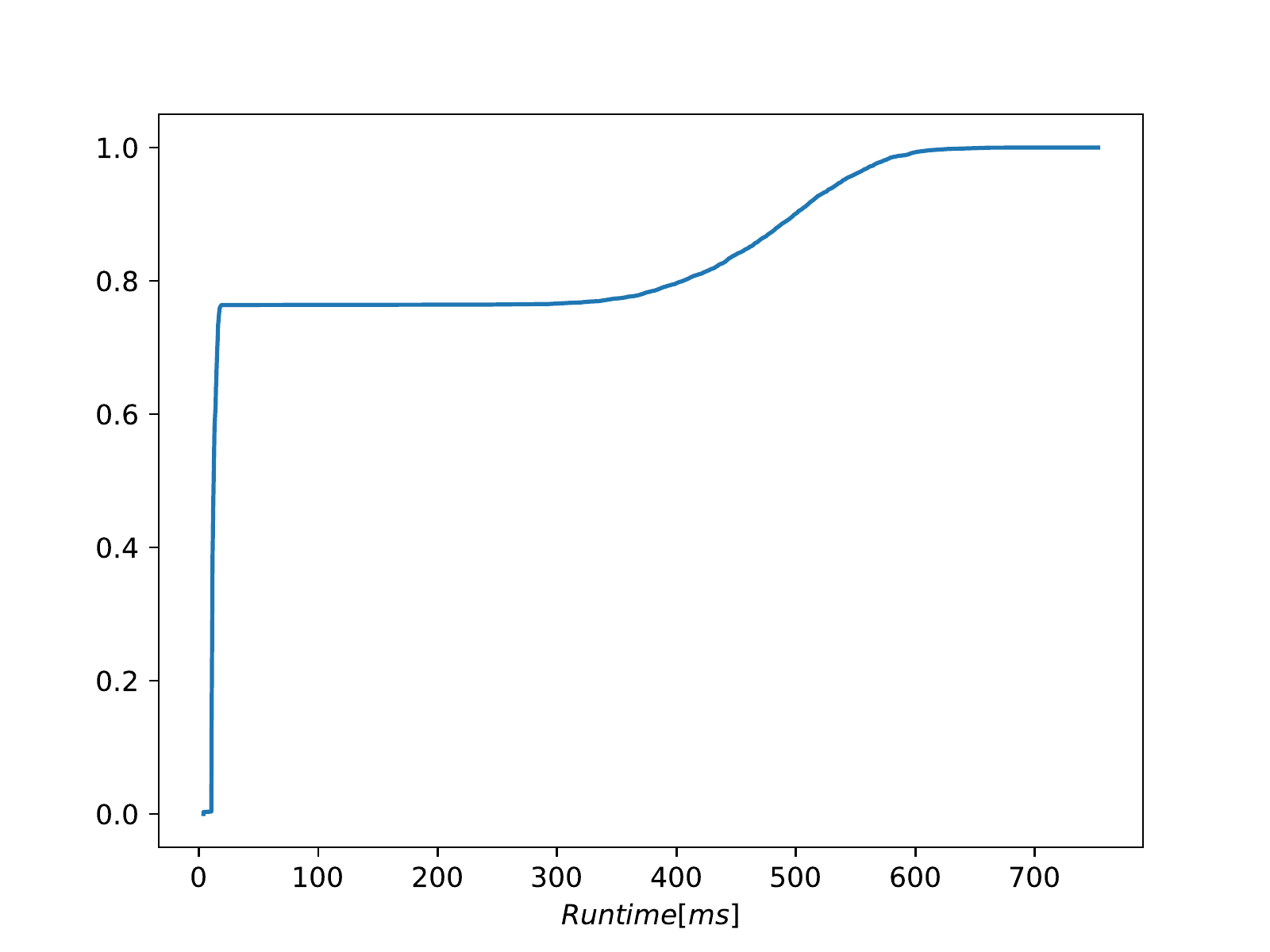}}\\
\subfigure[cifar\_crown\_large\_8\_255]{\includegraphics[width=0.5\columnwidth]{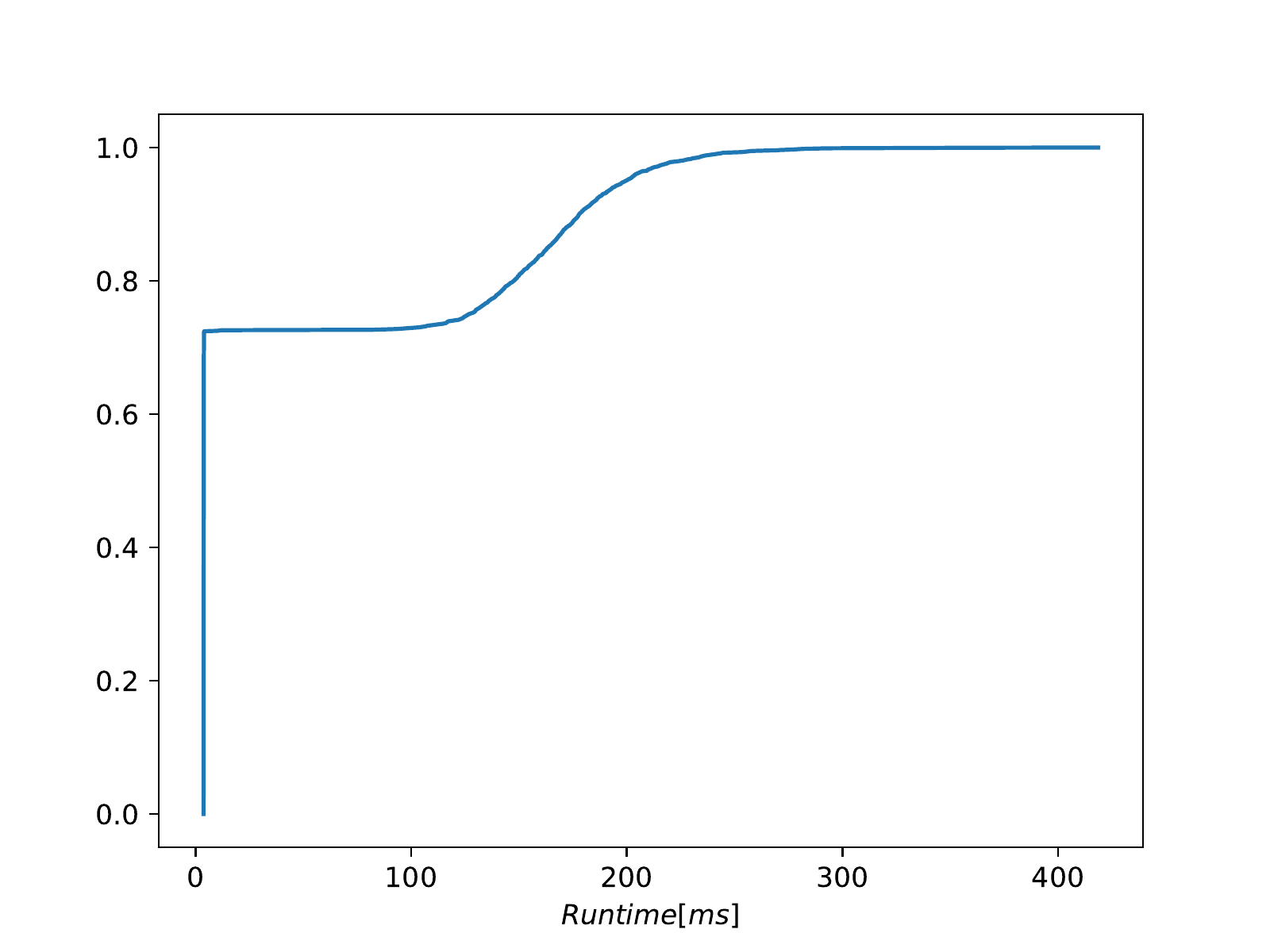}}
\subfigure[convSuper\_mnist]{\includegraphics[width=0.5\columnwidth]{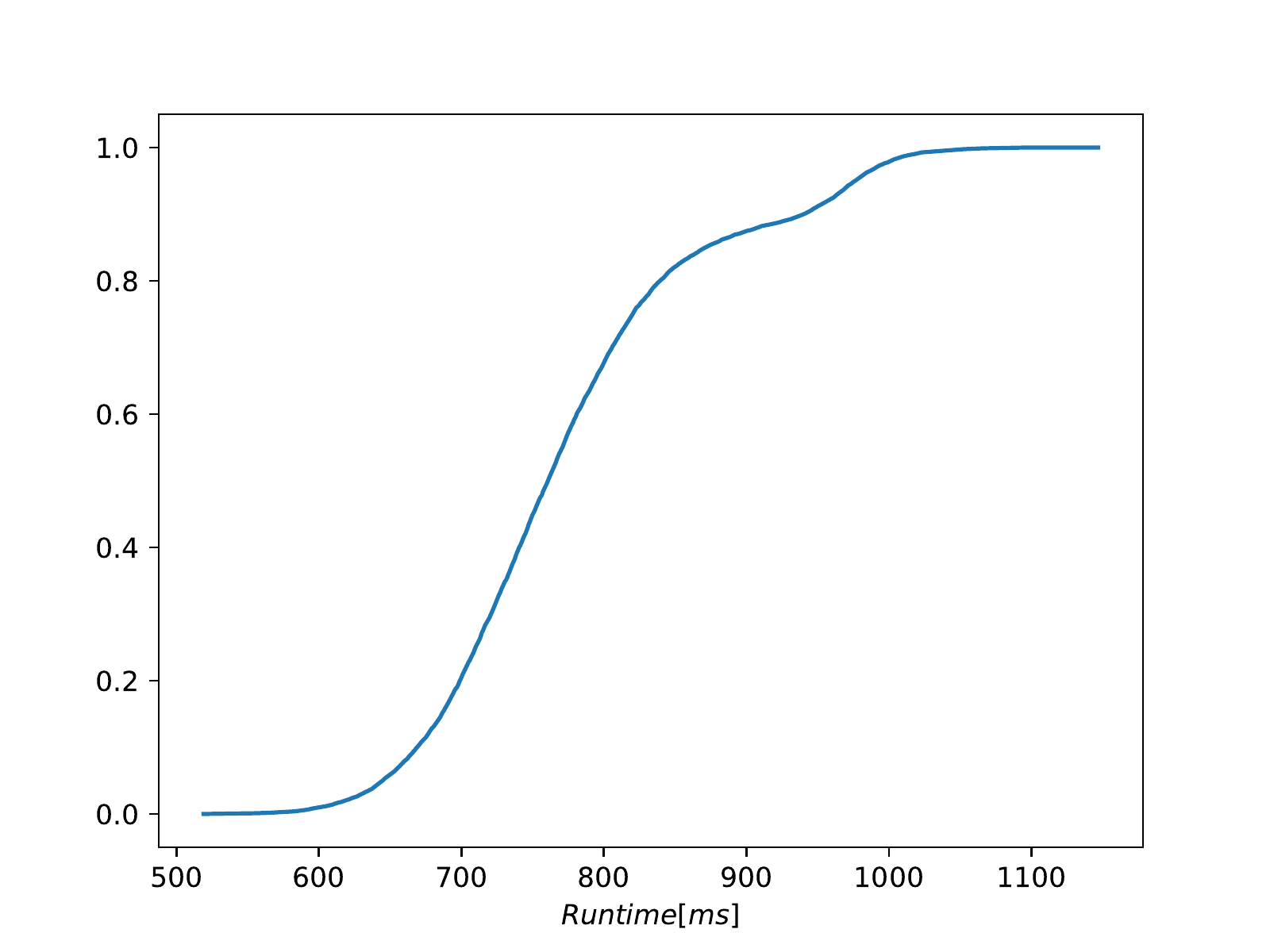}}
\subfigure[mnist\_crown\_large\_0.2]{\includegraphics[width=0.5\columnwidth]{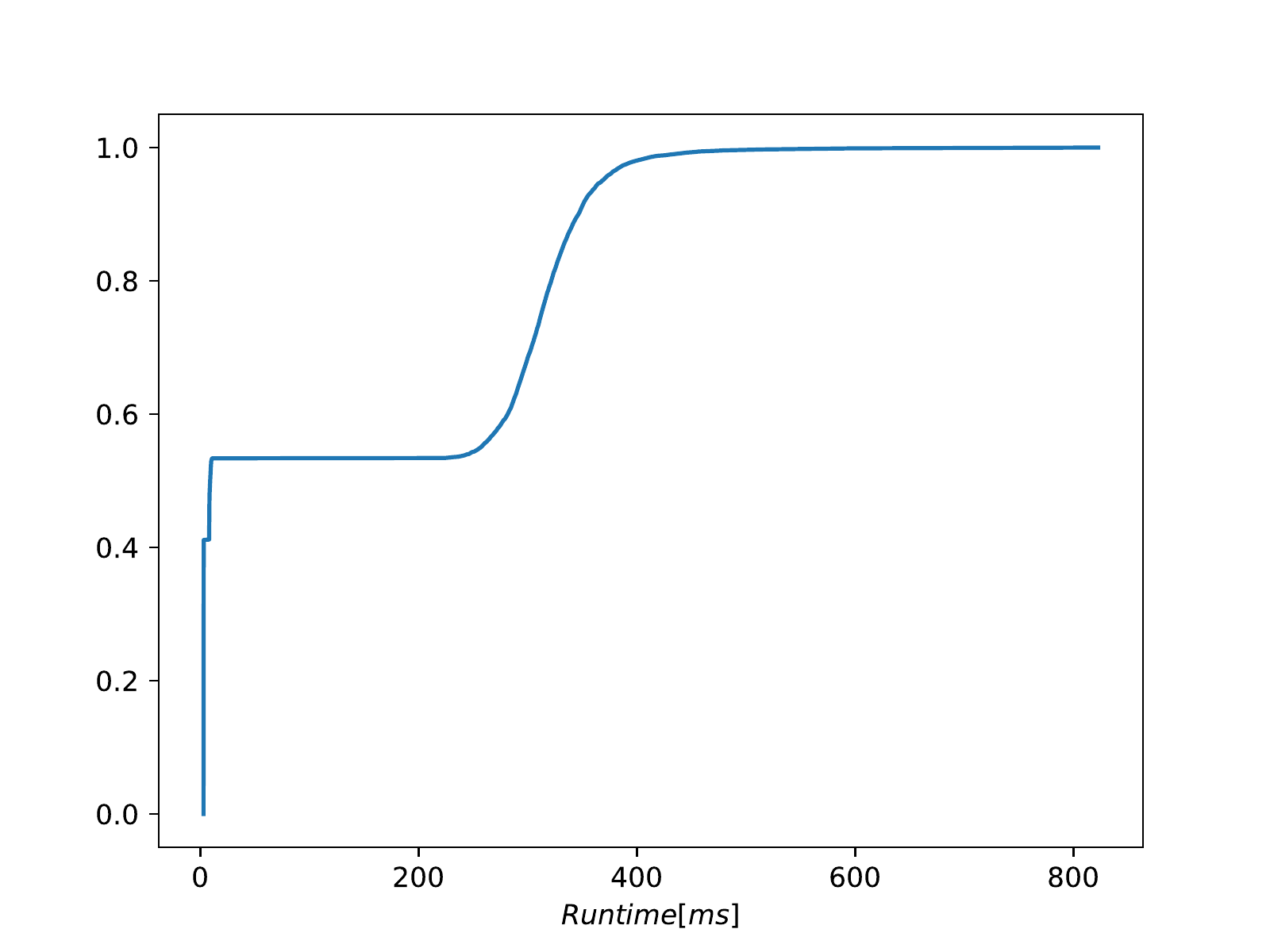}}
\subfigure[mnist\_crown\_large\_0.4]{\includegraphics[width=0.5\columnwidth]{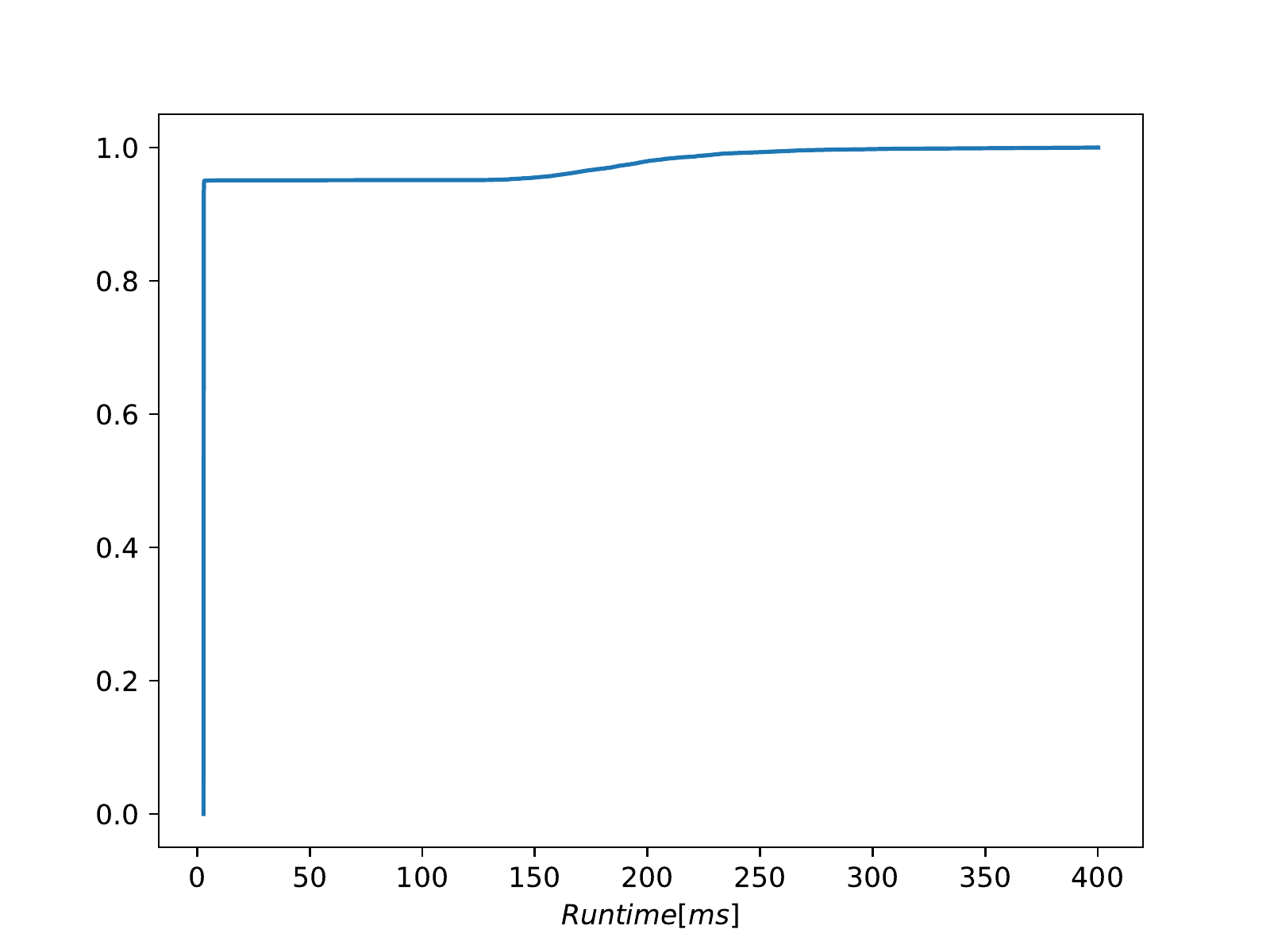}}
\caption{CDF plot of the runtime of \tool on the networks shown in Table~\ref{Ta:networks}.}
\end{figure*}

Fig.~\ref{fig:cdf} shows the CDFs of the runtime of \tool for the different networks. While the runtimes are roughly normally distributed for normally and PGD trained networks, the CDFs of the runtimes for DiffAI and \crown trained networks have large tails on the right side. The reason for this is that most of the time early termination will result in a very low runtime for robustly trained networks, but sometimes the runtime can be orders of magnitudes higher.


\end{document}